%% file: access.tex
\documentclass{ieeeaccess}
\usepackage{cite}
\usepackage{amsmath,amssymb,amsfonts}
\usepackage{algorithmic}
\usepackage{graphicx}
\usepackage{textcomp}
\def\BibTeX{{\rm B\kern-.05em{\sc i\kern-.025em b}\kern-.08em
    T\kern-.1667em\lower.7ex\hbox{E}\kern-.125emX}}

\usepackage[breaklinks=true,colorlinks,bookmarks=false]{hyperref}

\usepackage{url}
\usepackage[labelformat=simple]{subcaption}

\usepackage{multirow}
\usepackage{multicol}
\usepackage{diagbox}

\usepackage[percent]{overpic}

\newcommand{\R}{\mathbb{R}}
\newcommand{\cs}{\mathit{cs}}

\usepackage[table]{xcolor}

\usepackage{algorithm}

\def\modify#1{#1}
\def\reviewers#1#2#3{}
\def\modifybegin{}
\def\modifyend{}

\def\sreviewers#1#2#3{}

\begin{document}

\history{Received 28 August 2022, accepted 3 September 2022, date of publication 6 September 2022, date of current version 15 September 2022.}

\doi{10.1109/ACCESS.2022.3204755}

\year{2022}
\vol{10}

\title{Performance Evaluation of Action Recognition Models on Low Quality Videos}

\author{
    \uppercase{
        Aoi Otani\authorrefmark{1},
        Ryota Hashiguchi\authorrefmark{1},
        Kazuki Omi\authorrefmark{1}
    },
    \uppercase{
        Norishige Fukushima\authorrefmark{1}
    },
    \IEEEmembership{Member, IEEE}
    \uppercase{
        and Toru Tamaki\authorrefmark{1}
    },
    \IEEEmembership{Member, IEEE}
}

\address[1]{Nagoya Institute of Technology, Gokiso-cho, Showa-ku, Nagoya 466-8555, Japan}

\tfootnote{This work was supported in part by JSPS KAKENHI Grant Number JP22K12090.}

\markboth
{Otani \headeretal: Performance Evaluation of Action Recognition Models on Low Quality Videos}
{Otani \headeretal: Performance Evaluation of Action Recognition Models on Low Quality Videos}

\corresp{Corresponding author: Toru Tamaki (e-mail: tamaki.toru@nitech.ac.jp).}

\begin{abstract}
    \include{abstract}
\end{abstract}


\titlepgskip=-15pt

\maketitle

\input{main_text.tex}

\bibliographystyle{IEEEtran}
\bibliography{mybib}

\clearpage

\begin{IEEEbiography}[{\includegraphics[width=1in,height=1.25in,clip,keepaspectratio]{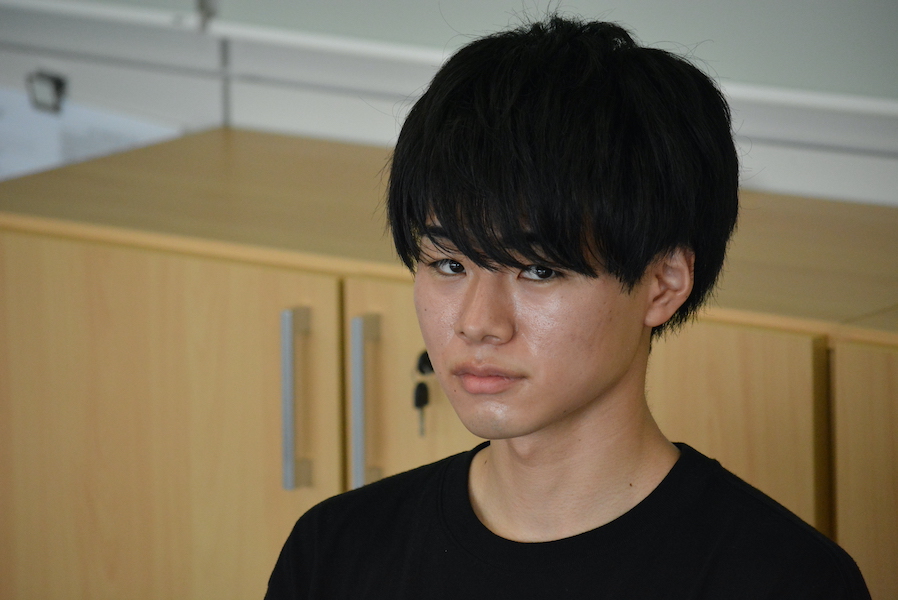}}]{Aoi Otani} received B.E. from Nagoya Institute of Technology in 2022.
    His research interests include computer vision and action recognition.
\end{IEEEbiography}

\begin{IEEEbiography}[{\includegraphics[width=1in,height=1.25in,clip,keepaspectratio]{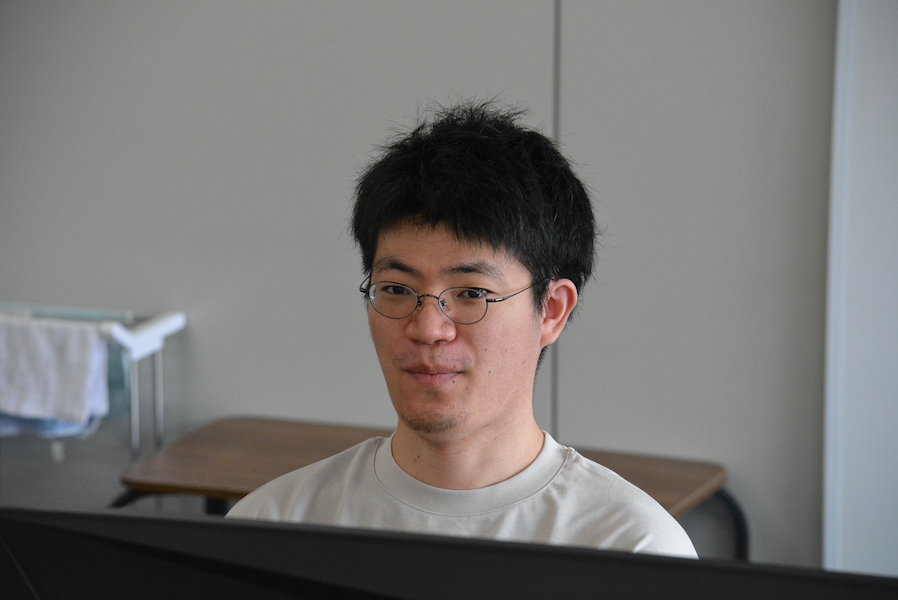}}]{Ryota Hashiguchi} received B.E. from Nagoya Institute of Technology in 2022.
    His research interests include computer vision and action recognition.
\end{IEEEbiography}

\begin{IEEEbiography}[{\includegraphics[width=1in,height=1.25in,clip,keepaspectratio]{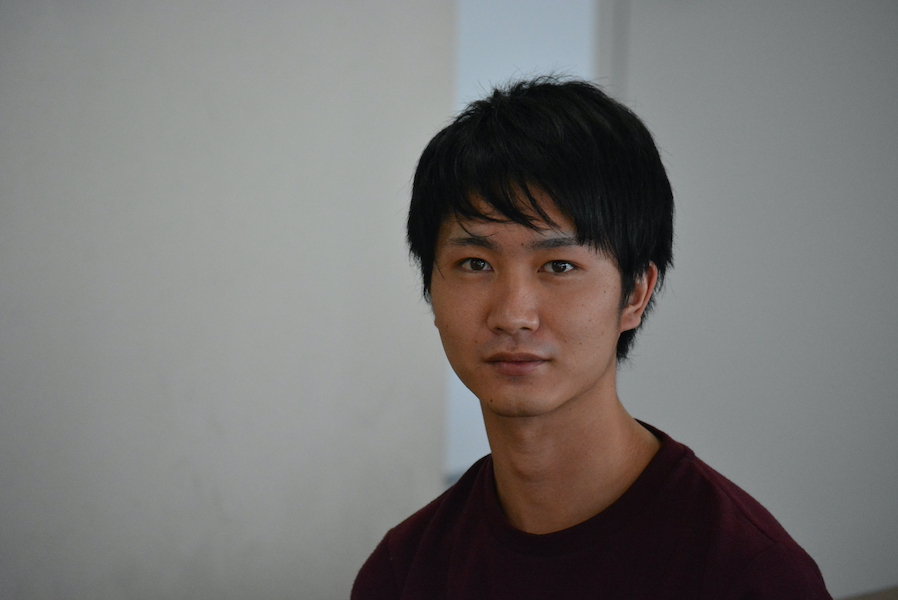}}]{Kazuki Omi} received B.E. from Nagoya Institute of Technology in 2022.
    His research interests include computer vision and action recognition.
\end{IEEEbiography}

\begin{IEEEbiography}[{\includegraphics[width=1in,height=1.25in,clip,keepaspectratio]{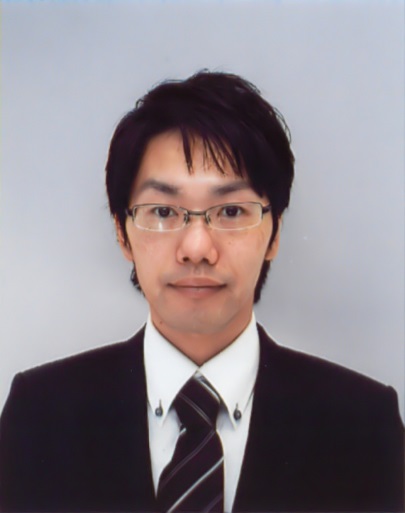}}]{Norishige Fukushima} received his B.E., M.E., and Ph.D. degrees from Nagoya University, Japan, in 2004, 2006, and 2009, respectively. He became an assistant professor in 2009 and associate professor in 2015 at Nagoya institute of technology, Japan. His research interests are image signal processing, parallel image processing, and compiler. He is a member of IEICE, IPSJ, and IEEE (CAS, SPS).
\end{IEEEbiography}

\begin{IEEEbiography}[{\includegraphics[width=1in,height=1.25in,clip,keepaspectratio]{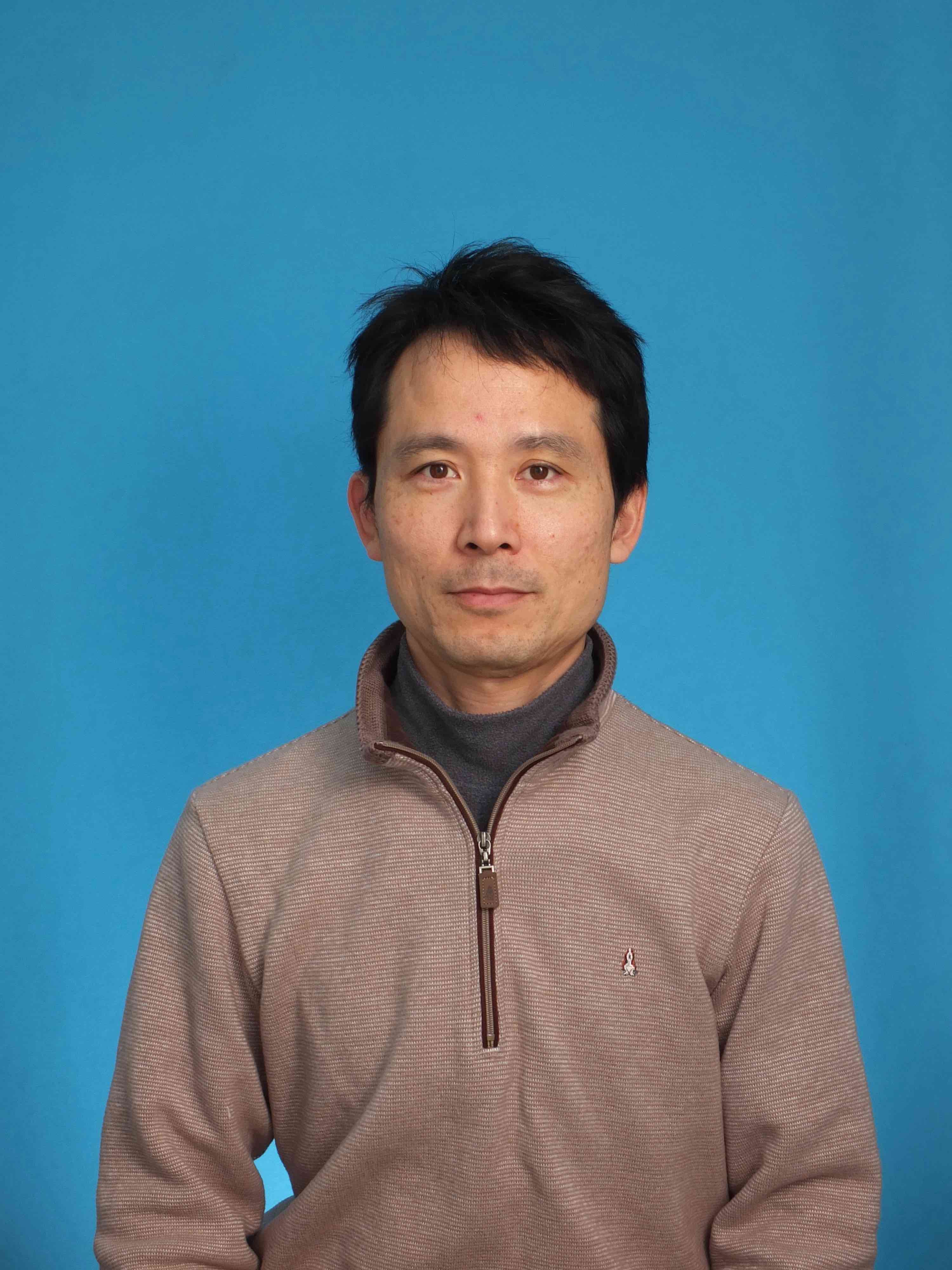}}]{Toru Tamaki} received his B.E., M.S., and Ph.D. degrees in information engineering from Nagoya University, Japan, in 1996, 1998 and 2001, respectively. After being an assistant professor at Niigata University, Japan, and an associate professor at Hiroshima University, Japan, he is currently a professor at the Department of Computer Science, Nagoya Institute of Technology, Japan. He was an associate researcher at ESIEE Paris, France, in 2015. His research interests include computer vision, image recognition, machine learning, and medical image analysis.
\end{IEEEbiography}

\EOD

\end{document}

%% file: abstract.tex

\modifybegin

%
In the design of action recognition models, the quality of videos is an important issue;
however, the trade-off between the quality and performance is often ignored.
In general, action recognition models are trained on high-quality videos,
hence it is not known
how the model performance degrades when tested on low-quality videos,
and how much the quality of training videos affects the performance.
The issue of video quality is important,
however, it has not been studied so far.
%
%
The goal of this study is
to show the trade-off between the performance and the quality of training and test videos
by quantitative performance evaluation of several action recognition models for transcoded videos in different qualities.
%
%
First, we show how the video quality affects the performance of pre-trained models.
We transcode the original validation videos of Kinetics400
by changing quality control parameters
of JPEG (compression strength) and H.264/AVC (CRF).
Then we use the transcoded videos
to validate the pre-trained models.
%
%
Second, we show how the models perform when trained on transcoded videos.
We transcode the original training videos of Kinetics400
by changing the quality parameters of JPEG and H.264/AVC.
Then we train the models on the transcoded training videos
and validate them with the original and transcoded validation videos.
%
%
Experimental results with JPEG transcoding show that there is no severe performance degradation
(up to $-1.5\%$) for compression strength smaller than 70 where no quality degradation is visually observed,
and for larger than 80 the performance degrades linearly with respect to the quality index.
Experiments with H.264/AVC transcoding show that
there is no significant performance loss (up to $-1\%$) with CRF30
while the total size of video files is reduced to $30\%$.
%
%
In summary, 
the video quality doesn't have a large impact on the performance of 
action recognition models unless the quality degradation is severe and visible.
This enables us to transcode the training and validation videos
and reduce the file sizes to one-third of the original videos.

\modifyend

%% file: main_text.tex
\section{Introduction}

In recent years, the technology to understand human action through video analysis by AI has been attracting attention~\cite{Review,survey1,survey2,Video_Transformers_Survey}.
For this purpose, it is necessary to recognize the actions of people in videos, such as walking or running, and such a task is called \textit{action recognition}.
With the advent of large datasets~\cite{Kinetics_400,AVA,YouTube-8M,ActivityNet,hvu,Moments_in_Time,ssv1_v2,UCF101,HMDB51} and the development of deep learning technologies, tremendous action recognition methods have been proposed~\cite{X3D,SlowFast,hara3dcnns,TimeSformer,video_swin_transformer}.
These technologies are now being used in a wide range of fields: detection of abnormal behaviors, 
skill assessment of workers, 
and support for sports training. 

Action recognition requires understanding various situations, such as interactions between people and objects in a scene.
However, there are many challenging problems in video action recognition, and also the video content is diverse that contains various problems.
For example, the following factors make action recognition difficult:
people size 
and appearance, 
viewpoint, 
occlusions, 
lighting, 
shadows, 
and data size.

In this study, we investigate the effect of video encoding quality on action recognition performance among those problems.
Existing action recognition models are not designed for low-quality videos; thus, it is not clear how they perform on such low-quality videos.
There are the following two potential situations where low-quality videos are used.

First, when we consider actual applications of action recognition (Fig.~\ref{fig:original}), the input videos might not be high-quality.
The videos provided by action recognition datasets commonly used in research today are usually high-resolution and high-quality.
However, the resolution of the image might be small if a camera captures the action scene from a distance and the region of interest (ROI) is cropped (Fig.~\ref{fig:low_resolution}).
In addition, if the camera transmits the video through a connection with limited bandwidth, the video quality will be considerably degraded (Fig.~\ref{fig:low_quality}).
This situation also occurs in streaming, where the bit rate is dynamically adjusted according to communication conditions.
In recent years, using pre-trained models has become common in many applications, but it is not clear how they perform on such low-quality videos.

\input{fig/initial_img_compare}

Second, the size of the video datasets has been increasing in recent years, and the required disk space has also been increasing.
When we train action recognition models, it is a common choice to store frames of each video as a sequence of numbered JPEG images in advance and load the JPEG images during training.
Moreover, a common pre-processing for training is to resize the shorter side of the image in the range of 256 and 320 pixels while maintaining the aspect ratio and then randomly crop a patch of $224 \times 224$ pixels.
Therefore, videos (or JPEG images) are usually resized to save space.
If video or image files are further transcoded with a high compression factor, disk space can be further reduced, and the speed of loading video and image files becomes faster.
The speed factor is essential because the smaller the size of files, the faster the files can be loaded via a local network or even over the cloud.
However, such high compression transcoding is usually not performed since it causes quality degradation and may affect the model performance.

There is still an open question about how action recognition models are affected by the quality of the input videos.
For image recognition (not action recognition), there are various studies on the effect of image quality, including object and face recognition at low resolution~\cite{Minimal_Images,Object_Recognition,Very_Low_Resolution,Fine_Grained}, investigations of the effect of recognition models on various image quality degradation such as blur, noise, contrast change, and JPEG compression~\cite{benchmarking,Understanding_How}.
For action recognition, several methods for low-resolution videos have been proposed~\cite{Extreme_Low_Resolution,TinyVIRAT}; however, the approaches modulate the quality degradation problem by image enhancement to ignore the effect of the video quality degradation.

Therefore, in this study, we quantitatively evaluate the trade-off between the video quality and the performance of action recognition models on Kinetics400~\cite{Kinetics_400}, an action dataset commonly used for evaluating model performance.
We show two types of experimental studies; one is validating pre-trained models on degraded videos of the validation set, and the other is training models on degraded videos of the training set.
We mainly focus on the quality degradation associated with transcoding by JPEG, a still image encoder, and H.264/AVC, a video encoder.
More artifacts appear when these codecs transcode images and videos with a higher compression ratio.

\begin{itemize}

\item
For JPEG transcoding, we store each input video frame as a sequence of numbered JPEG images (with a fixed frame rate at 30 fps).
The quantization parameter is specified to change the compression ratio.
To facilitate experiments of validating pre-trained models, we use data augmentation~\cite{imgaug} to simulate the JPEG compression; that is, we read original video files and apply JPEG compression to each frame, varying the quality parameter in a wide range.
We actually transcode video frames to JPEG files with a limited range of quality parameters in experiments for training models.

\item 
For H.264/AVC transcoding, we re-encode input videos at a reduced bit rate.
We use H.264/AVC codec for transcoding because datasets such as Kinetics400, which is commonly used for action recognition, are YouTube-based, and downloaded videos are usually encoded in H.264/AVC.

\end{itemize}

We explain the details of these two transcode settings in section~\ref{sec:transcoding}.
In section~\ref{sec:evaluation}, we evaluate the performance of well-known pre-trained models on validation videos transcoded by JPEG and H.264/AVC.
Then in section~\ref{sec:training} we show how transcoding affects the performance of models trained on transcoded training videos.

\section{Related Work}

\noindent\textbf{Recognition of low-resolution images.}
The resolution of the region to be recognized is substantially small in the following conditions: images taken under in-the-wild conditions and the images taken from a distance.
A simple approach is to convert a low-resolution image to a high-resolution image using super-resolution and then apply existing recognition methods~\cite{Fine_Grained,Dual}.
Other methods have been proposed that use class-specific domain knowledge to cope with low-resolution~\cite{DeriveNet} and simultaneously use high-quality images~\cite{Face_Recognition}.

\noindent\textbf{Analysis of image quality degradation.}
When the image quality is degraded, the recognition performance may be significantly degraded in applications using pre-trained CNN models.
Some works quantitatively evaluated how well general CNN models perform against various transformations to images (noise, blurring, compression, etc.)~\cite{Understanding_How,Why_do_deep,how_image_degradations,strengths_and_weaknesses}.
For quantitative evaluation, ImageNet-C/P~\cite{benchmarking}, a dataset that applies various corruptions and perturbations to ImageNet~\cite{ImageNet}, has also been proposed.
Various methods have been proposed to deal with this problem, including deep degradation prior~\cite{deep_degradation_cvpr_2020} and fine-tuning after learning reconstructed images from degraded ones~\cite{enhance_visual_recognition_ieee_2019}.

\noindent\textbf{Action recognition of low-resolution videos.}
Super-resolution for low-resolution videos is a solution to recognize low-resolution videos~\cite{Extreme_Low_Resolution,TinyVIRAT}.
In addition, there is a method dealing with the low-frame rate~\cite{on_the_effect}, which is slower than a typical frame rate of 30 fps.
However, there are \modify{not many} works on performance degradation for low-quality videos so far, except Video Enhancement Network~\cite{compressed_video}.
The Video Enhancement Network converts compressed low-quality video frames to high-quality video frames, using a pixel-level loss between input and output frames and also perceptual and adversarial losses.
Unlike this work, our study focuses on the experimental analysis of action recognition models
\modify{not tailored for} low-quality videos \modify{by showing the performance trade-off}.

\begin{figure}[t]

    \centering
    \includegraphics[width=.9 \linewidth]{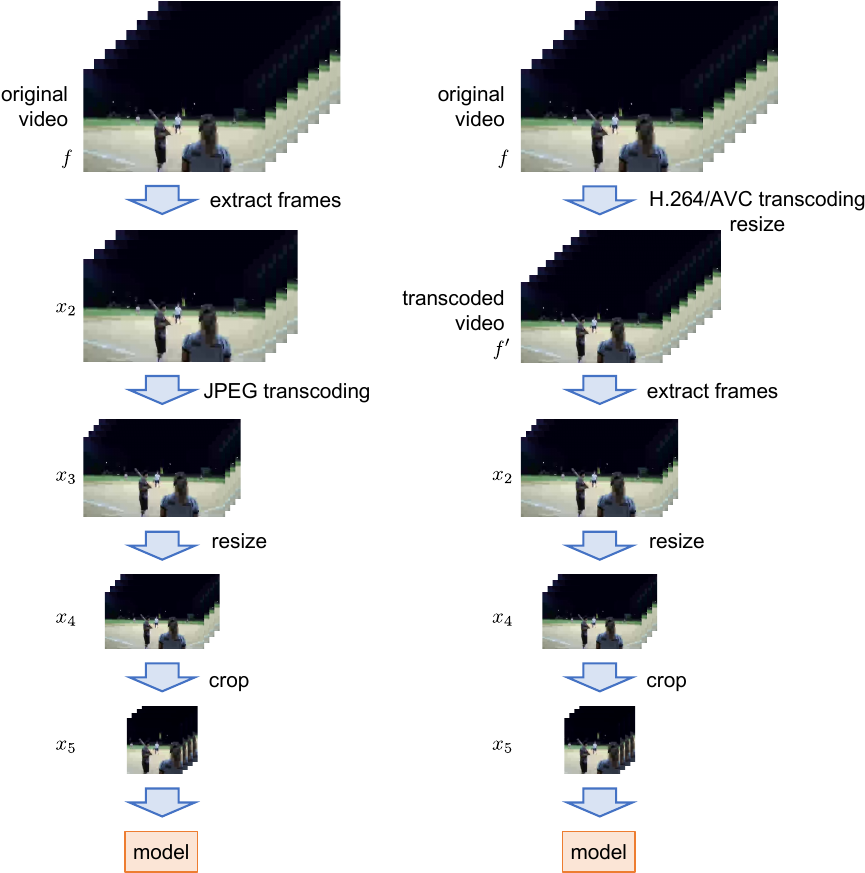}
    
    \medskip

    \caption{
    \modifybegin
    Overview of the procedures of
    (left) JPEG transcoding, and 
    (right) H.264/AVC transcoding.
    \modifyend
    }
    
    \label{fig:concept_figure}

\end{figure}

\input{fig/compare_images_0_100}

\input{fig/ffmpeg_frame}

\section{Transcoding Methods}
\label{sec:transcoding}

This section describes how videos are transcoded for experiments by JPEG and H.264/AVC.
\modify{%
The concepts of the transcoding procedures are shown in Figure~\ref{fig:concept_figure}.
}

\subsection{JPEG transcoding}

First, we describe how to simulate the JPEG transcoding of each input video frame.
In order to conduct the experiments in this study efficiently, we read each frame of a video and apply JPEG compression instead of storing frames of the video file as a sequence of JPEG image files.
For this purpose, we apply JPEG compression in \texttt{imgaug}~\cite{imgaug} as data augmentation with the parameter of compression strength ranging from 0 to 100.
The parameter is the inverse equivalent of the quality factor, with 100 for the highest compression.

Figure~\ref{fig:compare_images_10_100} shows the appearance of the compressed frames for the compression strength between 0 and 100.
From 0 to 80, there are no visible artifacts; however, block noise can be clearly seen, and significant image quality degradation occurs for the compression strength larger than 80.

\modifybegin
Algorithm~\ref{alg:jpeg_transcoding} shows
the procedure used to apply JPEG compression, resize, and crop video frames.
\modifyend

\begin{algorithm}[t]
\caption{
    \modify{Procedure of JPEG transcoding}
    }
\label{alg:jpeg_transcoding}

\begin{enumerate}

\item 
\label{jpeg_1}

Extract a video clip $x_1 \in \R^{sT_\mathrm{in} \times 3 \times H_\mathrm{in} \times W_\mathrm{in}}$ from an input video file $f$ using a data loader.
Here, $T_\mathrm{in}$ is the number of frames in a clip, $s$ is the stride between frames, and $H_\mathrm{in}$ and $W_\mathrm{in}$ are the height and width of video frames.
The frames are evenly sampled from $x_1$ with the stride of $s$, resulting in a clip $x_2 \in \R^{T_\mathrm{in} \times 3 \times H_\mathrm{in} \times W_\mathrm{in}}$.

\item
\label{jpeg_2}

Generate JPEG compressed version $x_3$ for each frame $x_2$ at the specified compression strength $\cs$.

\item
\label{jpeg_3}

\modifybegin
Resize frame $x_3$
so that the short side of the resized clip $x_4$ becomes the specified size $ss$ while maintaining the aspect ratio.
The size of $x_4$ is given as follows;
\begin{align}
    x_4 \in
    \begin{cases}
    \R^{T_\mathrm{in} \times 3 \times ss \times ss\frac{W_\mathrm{in}}{H_\mathrm{in}}} & H_\mathrm{in} \le W_\mathrm{in}\\
    \R^{T_\mathrm{in} \times 3 \times ss\frac{H_\mathrm{in}}{W_\mathrm{in}} \times ss} & H_\mathrm{in} \ge W_\mathrm{in}.
    \end{cases}
\end{align}
In other words, $x_4$ is resized to $ss$
regardless of the original size of $x_3$.
\modifyend

\item
\label{jpeg_4}

Crop the $H \times W$ pixel area in the center of frame $x_4$.
This results in a clip $x_5 \in \R^{T_\mathrm{in} \times 3 \times H \times W}$.
In our experiment, $H = W = 160$ or $H = W = 224$.

\end{enumerate}

\end{algorithm}

\subsection{H.264/AVC transcoding}
\label{sec:h264transcoding}

Next, we describe how to transcode input video files using H.264/AVC codec.
In this study, transcoding is performed by changing
two quality parameters: the group of pictures (GOP) size and the constant rate factor (CRF).
The experimental videos are transcoded by \texttt{FFmpeg}~\cite{ffmpeg} with x264%
\footnote{\url{https://www.videolan.org/developers/x264.html}}
\modify{which is a well-established H.264/ACV codec,}
and \texttt{-preset medium}%
\footnote{\url{https://trac.ffmpeg.org/wiki/Encode/H.264}}
for other parameters.

The larger the GOP size, the longer the interval between intra-frames (I-frames).
When the GOP size is 1, all frames are encoded as I-frames while the video file size becomes enormous.
The CRF values control video quality, and the valid range of CRF is between 0 and 51 (integer), with 0 being uncompressed and 51 being the maximum compression, and the default is 23.
Figure~\ref{fig:ffmpeg_frame} shows the changes in the quality of a transcoded video's frame by varying the GOP size and CRF.
No significant visual changes are observed at any GOP sizes up to CRF 40, but a noticeable quality decrease is observed at CRF 50.

\modifybegin
Algorithm \ref{alg:h264_transcoding} shows
the steps used to resize and transcode video frames.
\modifyend

\begin{algorithm}[t]
\caption{
    \modify{Procedure of H.264/AVC transcoding}
    }
\label{alg:h264_transcoding}

\begin{enumerate}

\item[$\cdot$]
(Preparation)
Transcode an input video file $f$ by H.264/AVC to create a video file $f'$ with specified GOP size and CRF.
\modifybegin
At the same time, frames are resized in a similar way as in step~\ref{jpeg_3} of Algorithm \ref{alg:jpeg_transcoding}, with $ss=360$ pixels (360p);
\begin{align}
    f' \in
    \begin{cases}
    \R^{T_\mathrm{in} \times 3 \times ss \times ss\frac{W_\mathrm{in}}{H_\mathrm{in}}} & H_\mathrm{in} \le W_\mathrm{in}\\
    \R^{T_\mathrm{in} \times 3 \times ss\frac{H_\mathrm{in}}{W_\mathrm{in}} \times ss} & H_\mathrm{in} \ge W_\mathrm{in}\\
    \R^{T_\mathrm{in} \times 3 \times H_\mathrm{in} \times W_\mathrm{in}}
    & \min(H_\mathrm{in}, W_\mathrm{in}) < ss,
    \end{cases}
\end{align}
where $T_\mathrm{in}$ is the number of frames in $f$,
$H_\mathrm{in}$ and $W_\mathrm{in}$ are the height and width of frames of $f$.
In other words, $f'$ is resized to $ss$ only when
the original size of $f$ is larger than $ss$.
\modifyend

\item 
Extract a video clip $x_2$ from the transcoded video file $f'$ using a data loader as in step~\ref{jpeg_1} of Algorithm \ref{alg:jpeg_transcoding}.

\item
Resize the frame as in step~\ref{jpeg_3} of Algorithm \ref{alg:jpeg_transcoding}.

\item
Crop the center area as in step~\ref{jpeg_4} of Algorithm \ref{alg:jpeg_transcoding}.

\end{enumerate}

\end{algorithm}

\begin{table}[t]

    \caption{
    Parameters for generating input video clips for each model.
    }
    \label{table:model_params}
    
    \centering
    
    \begin{tabular}{c|cccc}
                           & $ss$ & $W, H$ & $T_\mathrm{in}$ & $s$ \\ \hline
    \modify{X3D-XS}         & 181  & 160    & 4               & 12  \\
    \modify{X3D-S}          & 181  & 160    & 13              & 6   \\
    \modify{X3D-M}          & 256  & 224    & 16              & 5   \\
    \modify{SlowFast R50}   & 256  & 224    & 32              & 2   \\
    \modify{SlowFast R100}  & 256  & 224    & 32              & 2   \\
    \modify{3D ResNet R50}  & 256  & 224    & 8               & 8   \\
    \modify{TimeSformer}    & 256  & 224    & 8               & 32  \\
    \modify{Video Swin-B/T}   & 256  & 224    & 32              & 2  
    \end{tabular}

\end{table}

\section{Evaluation of pre-trained models}
\label{sec:evaluation}

In this section, we analyze the evaluations of the performance of common pre-trained models for action recognition with degraded videos of a validation set.

\subsection{Experimental Settings}
\label{sec:Experimental Settings for Evaluation}

\noindent\textbf{Dataset.}
We used Kinetics400~\cite{Kinetics_400}, the most common action recognition dataset consisting of a training set of 22k videos, a validation set of 18k videos, and a test set of 35k videos, with 400 categories of human actions.
Each video was collected from YouTube, and the action part was trimmed to 10 seconds in the video.
In this study, we used 19880 videos in the validation set%
\footnote{\url{https://github.com/cvdfoundation/kinetics-dataset}}
distributed for the Kinetics-700 Challenge 2021%
\footnote{\url{https://eval.ai/web/challenges/challenge-page/1054/overview}}.

\noindent\textbf{Models.}
We used the following recent state-of-the-art CNN models that are commonly used for comparison in the literature: \modify{X3D-M/S/XS~\cite{X3D}, SlowFast (R50/R101)~\cite{SlowFast}, and 3D ResNet (R50)~\cite{hara3dcnns},} as well as recent models based on Vision Transformer (ViT) \cite{survey3-vit}; \modify{TimeSformer~\cite{TimeSformer} and Video Swin Transformer (Video Swin-B)~\cite{video_swin_transformer}.}
These models were pre-trained on the original videos of the training set of Kinetics400.
\modifybegin
Pre-trained models of X3D, SlowFast, and 3D ResNet were obtained
from the PyTorchVideo~\cite{pytorchvideo} model zoo,
and models of TimeSformer and 
Video Swin were obtained from the official repositories.%
\footnote{%
\modify{%
Pre-trained models are available from the following repositories.\\
\url{https://pytorch.org/hub/facebookresearch_pytorchvideo_resnet/}\\
\url{https://pytorch.org/hub/facebookresearch_pytorchvideo_slowfast/}\\
\url{https://pytorch.org/hub/facebookresearch_pytorchvideo_x3d/}\\
\url{https://github.com/facebookresearch/TimeSformer}\\
\url{https://github.com/SwinTransformer/Video-Swin-Transformer} 
}
}
\footnote{%
\modify{%
We exclude any commercial models provided as APIs in commercial platforms as they are black-box to users, and changes to the underlying models and algorithms may occur unknowingly.
}
}
\modifyend
Table~\ref{table:model_params} shows the parameters to create input video clips for each model.
Note that the start frame of a clip is selected randomly in general, and models are evaluated by test time augmentation with multiple clips (this is called a multi-view test); however, in this experiment, the start frame was fixed to the beginning of the video to make evaluation reproducible with a single view.
\label{page:val_fix_frames}

\noindent
\modify{%
\textbf{Metrics.}
We report top-1 and top-5 for classification as these are common metrics
for action recognition \cite{SlowFast,X3D,hara3dcnns,TimeSformer,video_swin_transformer}.
}

\subsection{Results of JPEG transcoding}

\input{fig/acc_jpeg_compression}
\input{fig/acc_jpeg_image_score}

Figure~\ref{fig:acc_jpeg_compression_-10_100} shows the performance when the JPEG compression strength $\cs$ is increased by 10 from 0 to 100.
We can see that the trend of performance deterioration is similar for all models.
The performance does not change significantly up to $\cs=70$, but gradually decreases after 80, and at $\cs=100$, the recognition rate decreases to about half that of the original videos.
Figure~\ref{fig:acc_jpeg_compression_80_100} shows the performance for $\cs$ from 80 to 100 with the fine interval, and the performance gradually decreases as $\cs$ approaches 100.
As shown in Fig.~\ref{fig:compare_images_10_100}, there is no significant visual change in images transcoded with $\cs=80$ and $\cs=90$.
However, the model's performance deteriorates from $\cs=80$, and the performance drops about 10\% at $\cs=90$ compared to the original videos.

Next, we investigate the relationship between performance and the quality of transcoding results.
Figure~\ref{fig:acc_jpeg_compression_image_score} shows the performance against mean absolute error (MAE), root mean square error (RMSE), peak signal noise ratio (PSNR), and structure similarity (SSIM).
The results show a linear relationship between quality and performance; in other words, the performance of the models is proportional to the quality of the input video.
For MAE, RMSE, and SSIM, the slope $a$ and intercept $b$ of the linear approximation of the top-1 performance of X3D-M are $(a,b)=(-338, 76.4)$, $(-460, 77.6)$, $(221, -152.0)$ respectively.
We can say that the performance decreases by $338/256=1.32\%$
for each $1/256 \approx 0.004$ increase of MAE, and 1.80\% for RMSE (note that we divide by 256 since the range of pixel values in the input frame is $[0, 1]$ instead of $[0, 255]$).
Also, it can be seen that the performance degrades by 2.21\% for every 0.01 decrease in SSIM.

ViT-based models performed better than CNN-based models, as expected.
However, interestingly, Video Swin-B performs better than \modify{TimeSformer} at the worst quality ($cs > 95$).
This can also be seen in Fig.~\ref{fig:acc_jpeg_compression_image_score} that the slop of Video Swin-B is smaller than other models.
This indicates that the performance of ViT-based models can deteriorate as much as that of CNN-based models unless the model architecture is carefully designed, as in Video Swin-B.

\subsection{Results of H.264/AVC transcoding}

\input{fig/val_acc_ffmpeg}
\input{tables/acc_ffmpeg_tables}

Figure~\ref{fig:val_acc_ffmpeg} shows the performance of different models for transcoded videos by H.264/AVC of the validation set.
For all models, there is almost no performance difference when CRF is less than 20.
For larger CRF, we can see that the performance decreases more slowly with larger GOP sizes.
The worst performance is observed in the case of the GOP size of 1, in which case the videos have I-frames only.
For SlowFast-R101, there is a severe performance drop from CRF 40 to 50 with the GOP size of 2, compared to X3D-M and Slow-R50.
In the case of the transformer-based models, the performance deterioration due to the difference in GOP sizes (larger than 2) is not significant compared to the CNN-based models.

Table~\ref{tab:acc_ffmpeg_x3dm} shows how performances are degraded compared to the performance with the original videos.
When the GOP size is larger than 5, the performance drop is about 0.5\% to 1.0\% with CRF 30, but it becomes larger than 5\% with CRF 40 for all models.
In the case of CRF 50, the performance drops to 50\% of the original performance for CNN models and 70\% for ViT-based models.

\input{fig/accuracy_for_each_class_x3dm}

The results of Fig.~\ref{fig:val_acc_ffmpeg} show the average performance for all videos, and it is not clear whether the performance degrades in the same way for all action categories.
Figure~\ref{fig:accuracy_for_each_class_x3dm} shows the top-1 performance of X3D-M for each class with different CRF values (the GOP size was fixed to 12).
Since the recognition rate varies largely for different categories, this figure shows the relative performance to the reference (set to 100) of the original video (at CRF $-10$). For each CRF, 400 circles corresponding to different categories are plotted.
As can be seen from this figure, there is no significant performance degradation in all categories when CRF is less than 30.
Some categories show a significant performance decrease with CRF 40 (interestingly, some categories show an increase in performance), and almost all categories show a large performance drop with CRF 50.

\begin{table}[t]

    \definecolor{light_gray}{rgb}{0.9,0.9,0.9}
    \renewcommand{\g}{\cellcolor{light_gray}}

    \caption{
    Total file size (in GB) of transcoded 19880 validation videos of Kinetics400.
    Frames were resized to a maximum of 360 pixels on the short side.
    Shaded cells indicate that the file size has increased from 
    the total file size of the original videos of 30.56 GB.}
    \label{tab:dataset_crfg_filesize}
    \centering

    \begin{tabular}{c|cccccc}
    \diagbox{GOP}{CRF} & 0         & 10        & 20       & 30    & 40    & 50   \\ \hline
    1                  & \g 417.91 & \g 144.22 & \g 59.42 & 23.02 & 10.30 & 6.33 \\
    2                  & \g 327.40 & \g 137.67 & \g 58.12 & 23.97 & 10.26 & 5.81 \\
    5                  & \g 269.13 & \g 88.31  & \g 35.67 & 15.08 & 7.20  & 4.56 \\
    10                 & \g 249.66 & \g 74.47  & 28.80    & 12.08 & 6.13  & 4.14 \\
    12                 & \g 247.20 & \g 72.05  & 27.59    & 11.54 & 5.93  & 4.06
    \end{tabular}

\end{table}

The total file size of the videos transcoded with H.264/AVC is shown in Table~\ref{tab:dataset_crfg_filesize}.
In the case of the GOP size of 1 and CRF 0 (lossless), the file size (about 400 GB) becomes 13 times larger than the original videos (about 30 GB).
We can see that CRF must be 30 or higher for the benefit of transcoding the videos in terms of file size reduction.
In the case of CRF 30 with the GOP size of 12, the file size was reduced to about 40\% of the original video, with a performance loss of about 0.5\%.
Transcoding with CRF smaller than 20 does not change the performance, but the file size is the same or larger than the original videos, which is not beneficial.

\section{Training of action recognition models}
\label{sec:training}

In the previous section (Sec.~\ref{sec:evaluation}), we have shown the results of evaluating the performance of videos in the validation set transcoded by JPEG or H.264/AVC, using models that have been pre-trained on the \textit{original videos}.
Here, we show the performance of the models trained on the \textit{transcoded videos} in the training set.

We trained models with 240194 videos in the training set of the Kinetics-700 Challenge 2021 and evaluated models with the validation set, like the validation experiments for the pre-trained models.

The models used below were 
\modify{%
X3D-M~\cite{X3D}, 3D ResNet R50~\cite{hara3dcnns}, SlowFast R101~\cite{SlowFast},
TimeSformer~\cite{TimeSformer}, and Video Swin Transformer (Video Swin-T)~\cite{video_swin_transformer}
\footnote{
\modify{We used smaller Swin-T instead of Swin-B for efficient training.}
},
}
which have been pre-trained by Kinetics400%
\footnote{%
This is not usual because we used the same dataset for both pre-training and the downstream task.
However, we are interested in how the performance decreases from that obtained by pre-training.}.
\modifybegin
Adam~\cite{adam} was used for CNN-based models
with $\beta_1 = 0.9, \beta_2 = 0.999$ and learning rate of $10^{-4}$, and 
SGD was used for Transformer-based models
with the momentum of 0.9 and learning rate of $5 \cdot 10^{-4}$.
For both cases,
we used weight decay of $5 \cdot 10^{-5}$.
Parameters shown in Table~\ref{table:model_params} were used
to create input video clips for each model by randomly selecting the start frame of a clip for training.
\modifyend
Frame sampling of the input video clip to the model was performed as in the steps in Sec.~\ref{sec:h264transcoding}, except resizing the short side to a randomly chosen size in the range of [256, 320] pixels while maintaining the aspect ratio, followed by randomly cropping a $224 \times 224$ pixel patch and flipping it horizontally at a ratio of 50\%.
The batch size was 8, and training epochs were 5.

\modifybegin
For validation, the same procedure was used for creating validation clips as in Sec.~\ref{sec:Experimental Settings for Evaluation}.
\modifyend

\subsection{Results of JPEG transcoding}

\begin{table}[t]
    \caption{
    Total file size (in GB) of JPEG files transcoded from 240194 train videos and 19880 validation videos.
    Frames were resized to a maximum of 360 pixels on the short side.
    }
    \label{tab:dataset_ffmpeg_q_filesize_train_val}

    \centering
    
    \begin{tabular}{c|cccc}
    \texttt{-q}     & -1   & 10  & 17  & 40  \\
    $\cs$           & N/A  & 70  & 80  & 90  \\ \hline
    train file size & 3504 & 937 & 726 & 526 \\
    val file size   & 293  & 77  & 60  & 43 
    \end{tabular}

\end{table}

\begin{table}[t]
    \caption{Top-1 performance of models trained and evaluated on videos
     of Kinetics400 degraded by JPEG transcoding.}
    \label{tab:top-1_performance_dataset_ffmpeg_q_filesize_train_val}
    
    \centering

\modifybegin
    \begin{tabular}{c|cccc}
        \texttt{-q}   & -1    & 10    & 17    & 40  \\
        $\cs$         & N/A   & 70    & 80    & 90 \\ \hline
        X3D-M         & 59.09 & 58.90 & 59.12 & 57.59 \\
        3D ResNet R50 & 52.40 & 50.89 & 49.90 & 49.15 \\
        SlowFast R101 & 54.96 & 53.59 & 53.35 & 52.42 \\
        TimeSformer   & 76.60 & 76.05 & 75.55 & 74.20 \\
        Video Swin-T  & 68.43 & 67.45 & 66.93 & 65.33 \\
    \end{tabular}
\modifyend

\end{table}

In this experiment, we used sequences of numbered JPEG images transcoded from the original videos instead of applying data augmentation as in the evaluation experiments in the previous section.
We specified the quality factor for transcoding by using \texttt{FFmpeg}~\cite{ffmpeg} with a fixed frame rate of 30 fps.
We have seen that the performance deteriorates at the compression strength $\cs$ of 80 in the previous section; therefore, we chose three values 70, 80, and 90.
To align the value of $\cs$ and the quality scale \texttt{-q} of \texttt{FFmpeg} option, we used \texttt{-q} $=10, 17, 40$ based on a preliminary calibration (the value of \texttt{-q} ranges from $-1$ to 65, and the smaller the value, the better the quality).
Table~\ref{tab:dataset_ffmpeg_q_filesize_train_val} shows the total size of transcoded JPEG files with corresponding $\cs$ values.
As a reference, sizes with the maximum JPEG quality (i.e., set \texttt{-q} to $-1$) are also shown in the table, which is a usual way for dataset preparation.

Table~\ref{tab:top-1_performance_dataset_ffmpeg_q_filesize_train_val} shows the performance for different $\cs$ settings.
For X3D-M, the performance at $\cs=70$ is almost the same with that of the highest quality of JPEG transcoding, and the difference is only $0.1\%$.
Performances decreased with increasing $\cs$, with a drop of 3\% at $\cs=90$.
This result aligned with that in the validation experiments with pre-trained models.
For 3D ResNet and SlowFast, the performance deteriorates at $\cs=70$ by about 1.5\%, which is slightly worse than X3D-M; however, the results do not deviate significantly from the results of the validation experiment.

\subsection{Results of H.264/AVC transcoding}
\label{sec:training H.264/AVC}

For experiments with training videos transcoded by H.264/AVC, the performance was evaluated in the same way as in the previous experiments for the pre-trained models.
The 19880 videos in the validation set were transcoded with the corresponding transcoding setting of the training set.
For example, validation videos transcoded with CRF 30 were used for evaluating models trained on training videos transcoded with CRF 30.

Based on the results of evaluation experiments in the previous section, we fixed the GOP size to 12 and used CRF 30 as the standard compression setting, and varied CRF to 20, 30, and 40, as the setting of CRF 30 has a reasonable trade-off; a relatively small performance degradation but significant reduction of the file sizes.

Table~\ref{tab:train_model_val} shows the performance of models with different CRF settings.
The performance degrades about $1\%$ with CRF 30 compared to the original videos.
However, the performance of CRF 40 was decreased by more than 3\%, confirming the negative impact of overly transcoding on the recognition rate.

Table \ref{tab:dataset_crfg_filesize_train} shows the total file size of transcoded videos in the training set, and the transcoding with CRF 30 reduces the file size to about 40\%.
Therefore, it can be said that CRF 30 significantly reduces the file size while keeping the performance degradation to about 1\%.
This is consistent with the observation in the evaluation experiments of the pre-trained models.

\begin{table}[t]
    \caption{Top-1 performance for the transcoded validation set with models trained on the transcoded training set of Kinetics400.
    Both sets were transcoded by H.264/AVC with the same CRF setting.
    }
    \label{tab:train_model_val}
    \centering

\modifybegin
    \begin{tabular}{c|cccc}
    CRF           & original & 20    & 30    & 40    \\ \hline
    X3D-M         & 59.53    & 58.99 & 58.49 & 56.28 \\
    3D ResNet R50 & 51.29    & 52.63 & 50.86 & 47.05 \\
    SlowFast R101 & 54.58    & 54.85 & 53.58 & 50.44 \\
    TimeSformer   & 76.90    & 77.06 & 76.58 & 74.80 \\ 
    Video Swin-T  & 68.54 & 68.94 & 67.97 & 66.01 \\ 
    \end{tabular}
\modifyend

\end{table}

\begin{table}[t]
    \caption{
    Total file size (in GB) of 240194 train videos and 
    19880 validation videos of Kinetics400 transcoded by H.264/ACV.
    Frames were resized to a maximum of 360 pixels on the short side.
    The GOP length was fixed to 12.
    }
    \label{tab:dataset_crfg_filesize_train}
    \centering
    \begin{tabular}{c|cccc}
    CRF             & original & 20  & 30  & 40 \\ \hline
    train file size & 349      & 313 & 131 & 68 \\
    val file size   & 31       & 28  & 12  & 6 
    \end{tabular}
\end{table}

\modifybegin

\subsection{Results on HMDB51 with H.264/AVC}

In addition to the experiments on Kinetics400,
we used another dataset, HMDB51, to see if a similar trend is observed for training with
H.264/AVC transcoding.

HMDB51~\cite{HMDB51} consists of 3.6k training videos and 1.5k validation videos with 51 human action categories.
Each video was collected from different sources such as movies, the web, and YouTube. The shortest video is less than 1 second, and the longest is about 35 seconds, while most videos are between 1 and 5 seconds long, with an average length of 3.15 seconds.
The short side of frames was resized to 240 pixels.
The first split was used in this experiment as it is common to be reported.
Training and validation procedures were the same with experiments on Kinetics400 (Sec.~\ref{sec:training H.264/AVC}), except that 
the last linear layer of each model was replaced for fine-tuning with random initialization, and 
the training epochs were set to 50 for a reasonable convergence.

Table \ref{tab:train_h264_hmdb51} shows the performance of models with different CRF settings,
like as Table \ref{tab:train_model_val} for Kinetics400.
The results clearly show a similar trend;
transcoding with CRF 30 did not show
significant performance loss (about $1\%$)
while CRF 40 decreased the performance by $2\% \sim 4\%$.

Table~\ref{tab:dataset_crfg_filesize_hmdb51} shows the total file size of transcoded videos of HMDB51, and the transcoding with CRF 30 reduces the file size to about 22\%, which is smaller than 30\% for Kinetics400 shown in Table \ref{tab:dataset_crfg_filesize_train}.

\begin{table}[t]
    \caption{
    \modify{
    Top-1 performance for the transcoded validation set with models trained on the transcoded training set of HMDB51.
    Both sets were transcoded by H.264/AVC with the same CRF setting.
    }
    }
    \label{tab:train_h264_hmdb51}
    \centering

\modifybegin
    \begin{tabular}{c|cccc}
    CRF           & original & 20    & 30    & 40    \\ \hline
    X3D-M         & 69.24    & 70.22 & 68.00 & 64.99 \\
    3D ResNet R50 & 49.08    & 48.17 & 48.10 & 45.68 \\
    SlowFast R101 & 59.22    & 64.20 & 60.54 & 58.97 \\
    TimeSformer   & 71.73    & 71.99 & 71.66 & 67.93 \\ 
    Video Swin-T  & 74.21    & 75.52 & 75.33 & 70.22 \\ 
    \end{tabular}
\modifyend

\end{table}

\begin{table}[t]
    \caption{
    \modify{
    Total file size (in MB) of train and 
    validation videos of HMDB51 transcoded by H.264/ACV.
    The GOP length was fixed to 12.
    }
    }
    \label{tab:dataset_crfg_filesize_hmdb51}
    \centering

\modifybegin
    \begin{tabular}{c|cccc}
    CRF             & original & 20  & 30  & 40  \\ \hline
    train file size & 1153     & 852 & 265 & 101 \\
    test  file size & 472      & 346 & 108 & 42
    \end{tabular}
\modifyend

\end{table}

\modifyend

\section{Discussions and limitations}

\modify{
Although we have done the experiments above with video files, similar results would be obtained for video streams}
transmitted over poor network connections, as mentioned in the introduction.
Figure~\ref{fig:val_acc_ffmpeg} shows how the performance degrades with reduced bit rates.
The original video files have bit rates of about 1000 kb/s (the resolution is up to 360p), and the pre-trained models keep the performance with a slight degradation even for 10 times lower bit rates.
This result may be the key to guaranteeing the performance when deploying models trained on high-quality video files against video streams that are transmitted from edge devices and processed on the cloud.

In terms of the file size, the results could suggest a good cost-performance trade-off.
It might not be cheap to store a large video dataset on cloud storage when the training is performed on the cloud, but the size can be reduced in trade for some performance degradation.
Using JPEG files may be a current common approach to training action recognition models in academic research, which are often performed on physical servers.
In this case, the total size of the highest quality JPEG file is more than 3.5 TB, as shown in Table ~\ref{tab:dataset_ffmpeg_q_filesize_train_val}, which might just barely fit on a slow SATA3-SSD, but not on a faster NVMe-SSD (the current mainstream size is smaller than 2 TB).
Our results indicate how much performance degrades when the dataset is transcoded to fit in specified storage size.

\modifybegin

However, to find the trade-off, the computation cost is high for evaluating different settings of transcoding. In experiments, we used 30 different values of compression strength for JPEG transcoding and 30 more different settings of GOP and CRF for H.264/AVC transcoding. 
Also, parameters and encoders of transcoding are diverse, and all of them were not covered. The results were therefore limited to the current experimental setting.

\modifyend

\section{Conclusions}

In this paper, we have analyzed quantitative evaluations of the relationship between video quality and the performance of action recognition models through experiments using Kinetics400.
The experiments with JPEG transcoding showed no severe performance degradation when compression strength $\cs$ is below 70, which is the range where no quality degradation is visually observed, and the performance degraded linearly with respect to the quality index when $\cs$ is above 80.
The experiments with H.264/AVC transcoding showed that there was no significant performance loss (up to $-1\%$) even with CRF 30, which can reduce the file sizes by about 30\%.
The same results were observed for the evaluation with the pre-trained model and with the models trained on transcoded videos.

Future work includes investigating whether similar trends can be observed using other large datasets of trimmed videos such as SSv2~\cite{ssv1_v2}, HVU~\cite{hvu} and Moments in Time~\cite{Moments_in_Time}, as well as untrimmed video datasets~\cite{ActivityNet},
\modify{
as well as the use of their recent encoders, such as H.265/HEVC. 
}

%% file: fig/initial_img_compare.tex
\begin{figure}[t]
    \centering
    \def\myimgwidth{0.3}
    \def\mybigimgscale{0.3}
    \def\mysmallimgsclae{3.054}

    \begin{tabular}{@{}c@{ }c@{ }c@{}}
        \begin{minipage}[t]{\myimgwidth\linewidth}
            \centering
            \includegraphics[width=\linewidth]{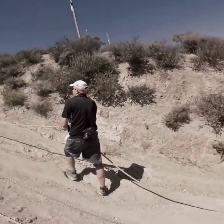} 
            \subcaption{}
            \label{fig:original}
        \end{minipage} &
        \begin{minipage}[t]{\myimgwidth\linewidth}
            \centering
            \includegraphics[width=\linewidth]{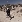}
            \subcaption{}
            \label{fig:low_resolution}
        \end{minipage} &
        \begin{minipage}[t]{\myimgwidth\linewidth}
            \centering
            \includegraphics[width=\linewidth]{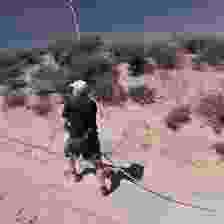} 
            \subcaption{}
            \label{fig:low_quality}
        \end{minipage} 
    \end{tabular}

    \caption{Degradation of image quality.
    (a) Original image.
    (b) Low resolution image.
    (c) Low quality image.
    }
    \label{fig:initial_img_compare}

\end{figure}

%% file: fig/compare_images_0_100.tex
\begin{figure*}[t]
    \centering
    \def\myimgwidth{0.1}
    \def\myimgsclae{0.37}
    
    \begin{tabular}{cccccc}
        \begin{minipage}[t]{\myimgwidth\hsize}
            \centering
            \includegraphics[width=\linewidth]{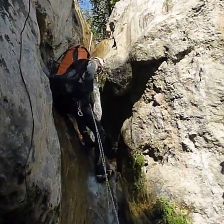}
            \subcaption{original}
            \label{fig:-10_100_original}
        \end{minipage} &
        \begin{minipage}[t]{\myimgwidth\hsize}
            \centering
            \includegraphics[width=\linewidth]{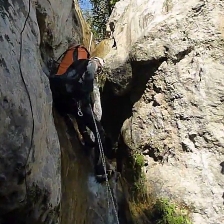}
            \subcaption{0}
            \label{fig:-10_100_0}
        \end{minipage} &
        \begin{minipage}[t]{\myimgwidth\hsize}
            \centering
            \includegraphics[width=\linewidth]{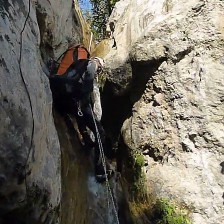}
            \subcaption{10}
            \label{fig:-10_100_10}
        \end{minipage} &
        \begin{minipage}[t]{\myimgwidth\hsize}
            \centering
            \includegraphics[width=\linewidth]{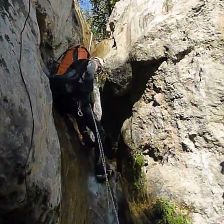}
            \subcaption{20}
            \label{fig:-10_100_20}
        \end{minipage} &
        \begin{minipage}[t]{\myimgwidth\hsize}
            \centering
            \includegraphics[width=\linewidth]{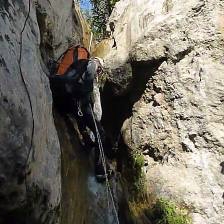}
            \subcaption{30}
            \label{fig:-10_100_30}
        \end{minipage} &
        \begin{minipage}[t]{\myimgwidth\hsize}
            \centering
            \includegraphics[width=\linewidth]{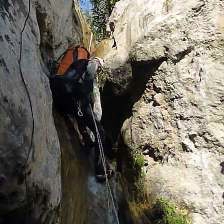}
            \subcaption{40}
            \label{fig:-10_100_40}
        \end{minipage} \\
        
        \begin{minipage}[t]{\myimgwidth\hsize}
            \centering
            \includegraphics[width=\linewidth]{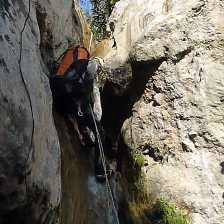}
            \subcaption{50}
            \label{fig:-10_100_50}
        \end{minipage} &
        \begin{minipage}[t]{\myimgwidth\hsize}
            \centering
            \includegraphics[width=\linewidth]{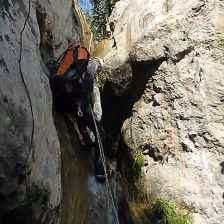}
            \subcaption{60}
            \label{fig:-10_100_60}
        \end{minipage} &
        \begin{minipage}[t]{\myimgwidth\hsize}
            \centering
            \includegraphics[width=\linewidth]{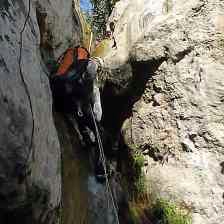}
            \subcaption{70}
            \label{fig:-10_100_70}
        \end{minipage} &
        \begin{minipage}[t]{\myimgwidth\hsize}
            \centering
            \includegraphics[width=\linewidth]{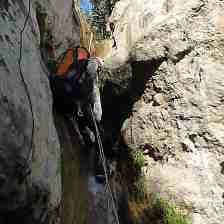}
            \subcaption{80}
            \label{fig:-10_100_80}
        \end{minipage} &
        \begin{minipage}[t]{\myimgwidth\hsize}
            \centering
            \includegraphics[width=\linewidth]{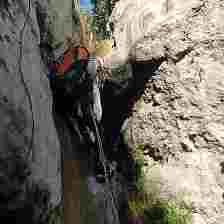}
            \subcaption{90}
            \label{fig:-10_100_90}
        \end{minipage} &
        \begin{minipage}[t]{\myimgwidth\hsize}
            \centering
            \includegraphics[width=\linewidth]{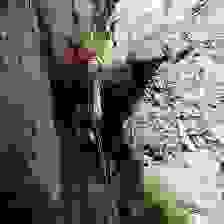}
            \subcaption{100}
            \label{fig:-10_100_100}
        \end{minipage} 
    \end{tabular}

    \caption{Changes of image quality due to JPEG transcoding
    with different values of compression strength
    (the higher the number, the more compressed the frame).}
    \label{fig:compare_images_10_100}
\end{figure*}

%% file: fig/ffmpeg_frame.tex
\begin{figure*}[t]

    \centering
    
    \scalebox{0.8}{

    \def\myimgwidth{0.1}
    \def\myimgsclae{0.11}
    

    \begin{tabular}{p{1.5cm}|cccccc}
        \diagbox[height=2\line, width=1.9cm]{GOP}{CRF} &  0 & 10 & 20 & 30 & 40 & 50 \\ \hline \\
        \multicolumn{1}{c|}{1} &
        \begin{minipage}[t]{\myimgwidth\hsize}
            \centering
            \includegraphics[width=\linewidth]{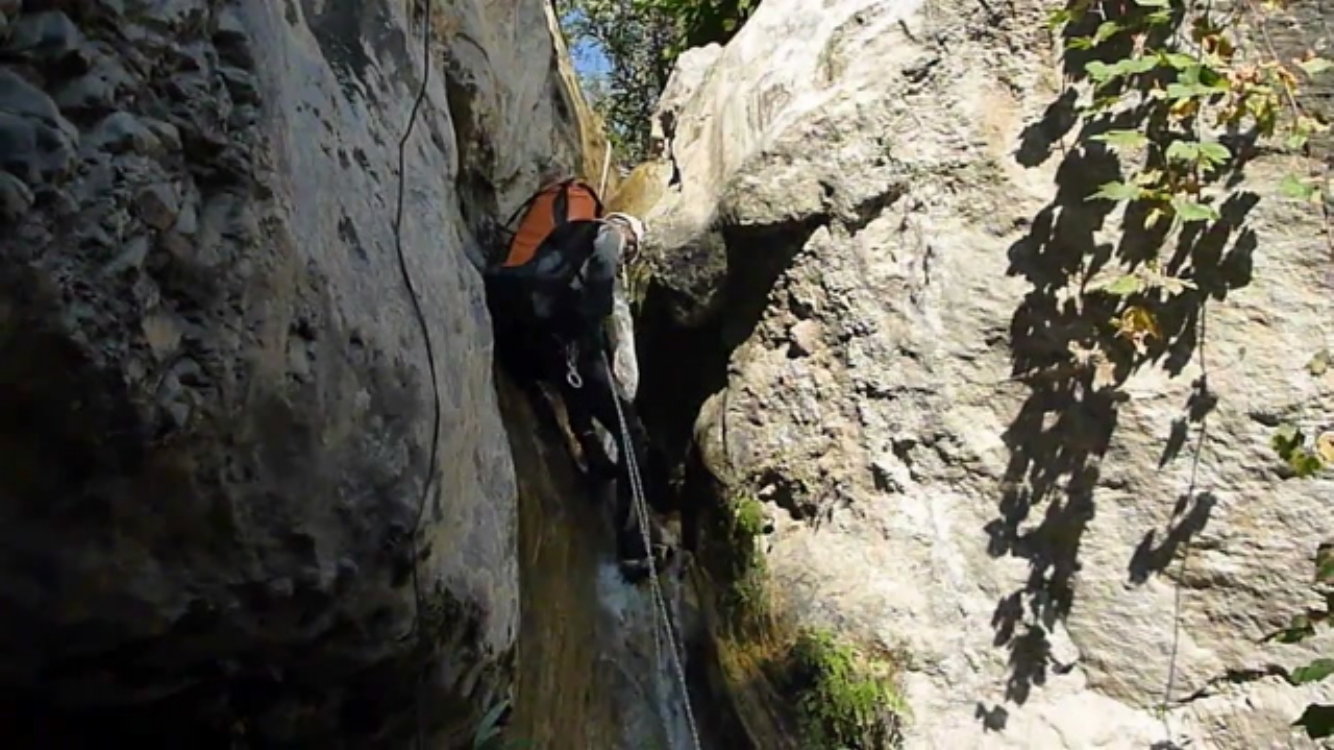}
            \label{fig:frame_gop1crf0}
        \end{minipage} &
        \begin{minipage}[t]{\myimgwidth\hsize}
            \centering
            \includegraphics[width=\linewidth]{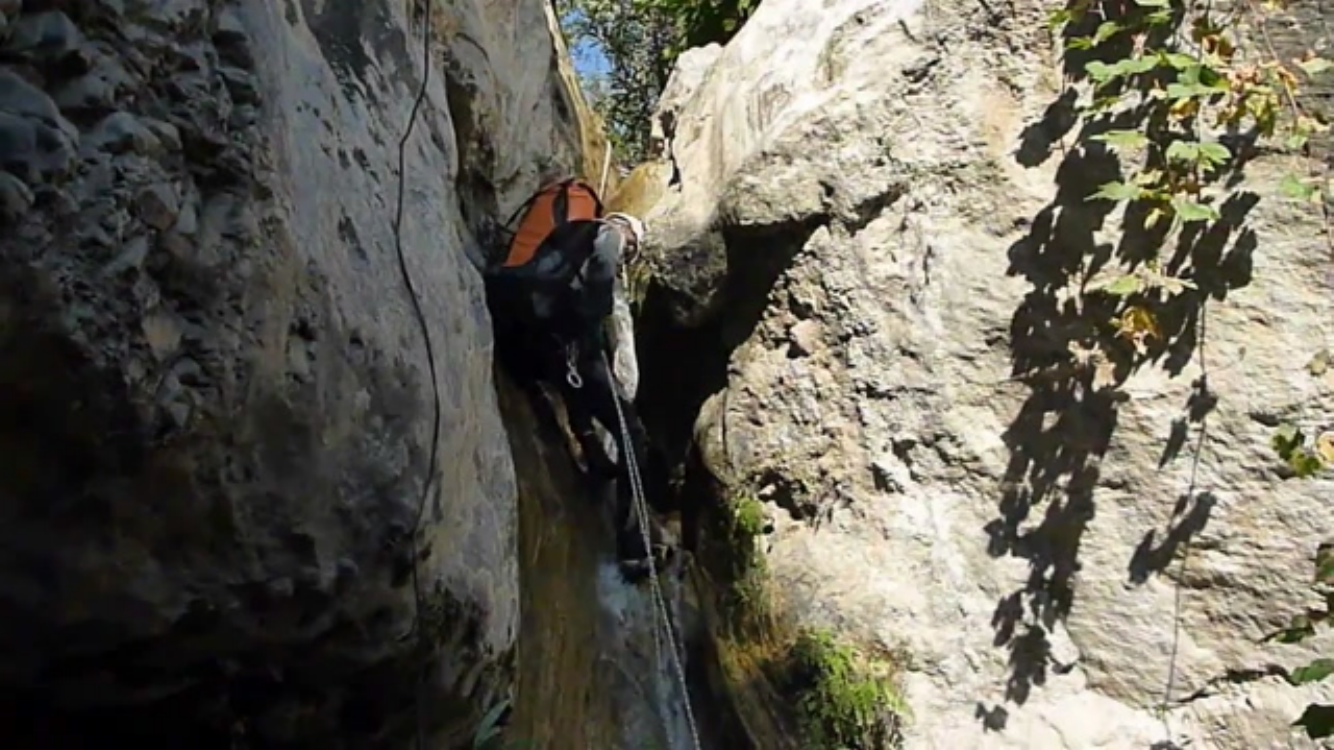}
            \label{fig:frame_gop1crf10}
        \end{minipage} &
        \begin{minipage}[t]{\myimgwidth\hsize}
            \centering
            \includegraphics[width=\linewidth]{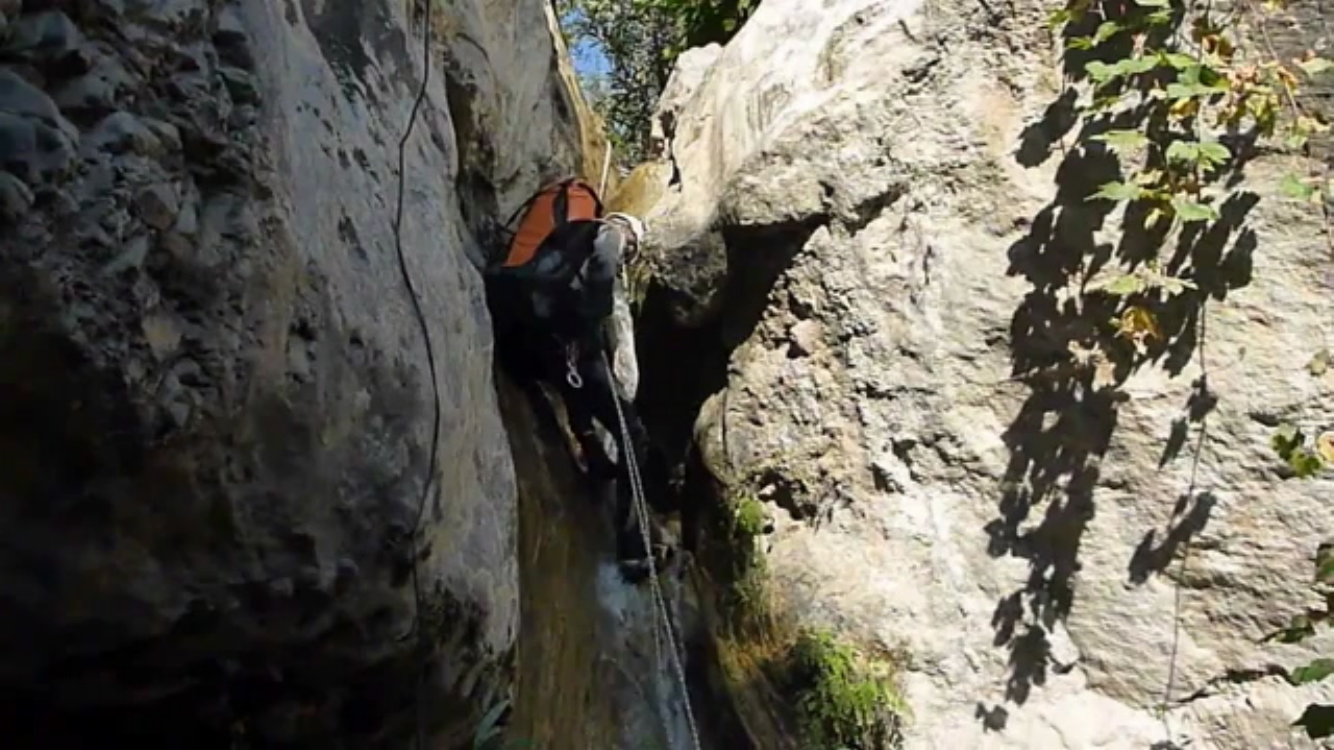}
            \label{fig:frame_gop1crf20}
        \end{minipage} &
        \begin{minipage}[t]{\myimgwidth\hsize}
            \centering
            \includegraphics[width=\linewidth]{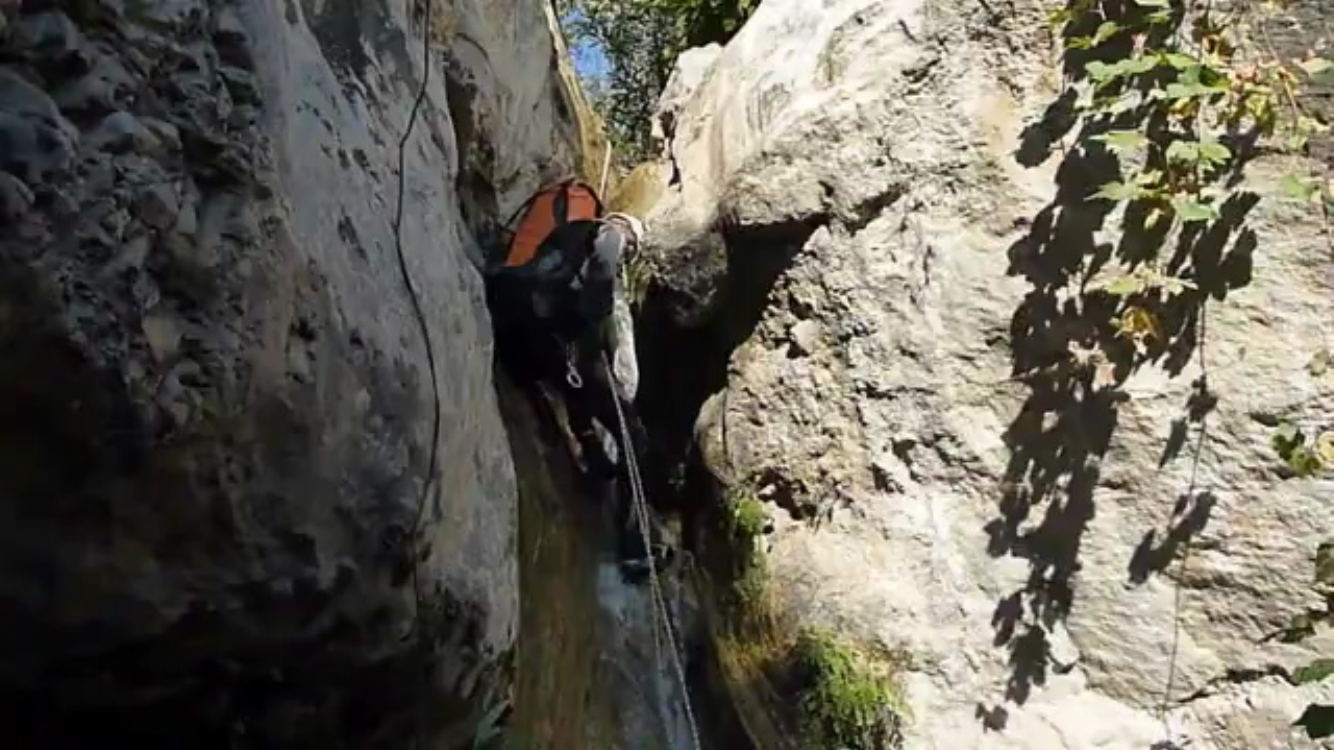}
            \label{fig:frame_gop1crf30}
        \end{minipage} &
        \begin{minipage}[t]{\myimgwidth\hsize}
            \centering
            \includegraphics[width=\linewidth]{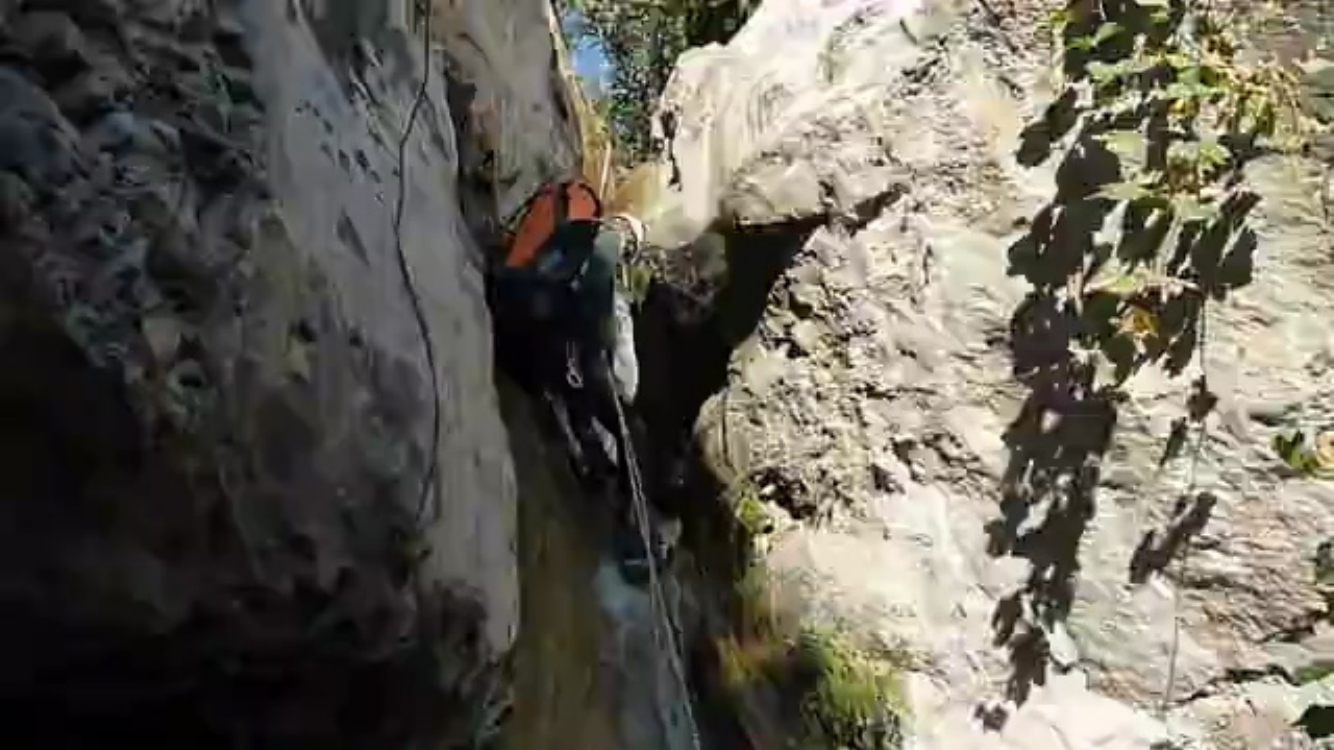}
            \label{fig:frame_gop1crf40}
        \end{minipage} &
        \begin{minipage}[t]{\myimgwidth\hsize}
            \centering
            \includegraphics[width=\linewidth]{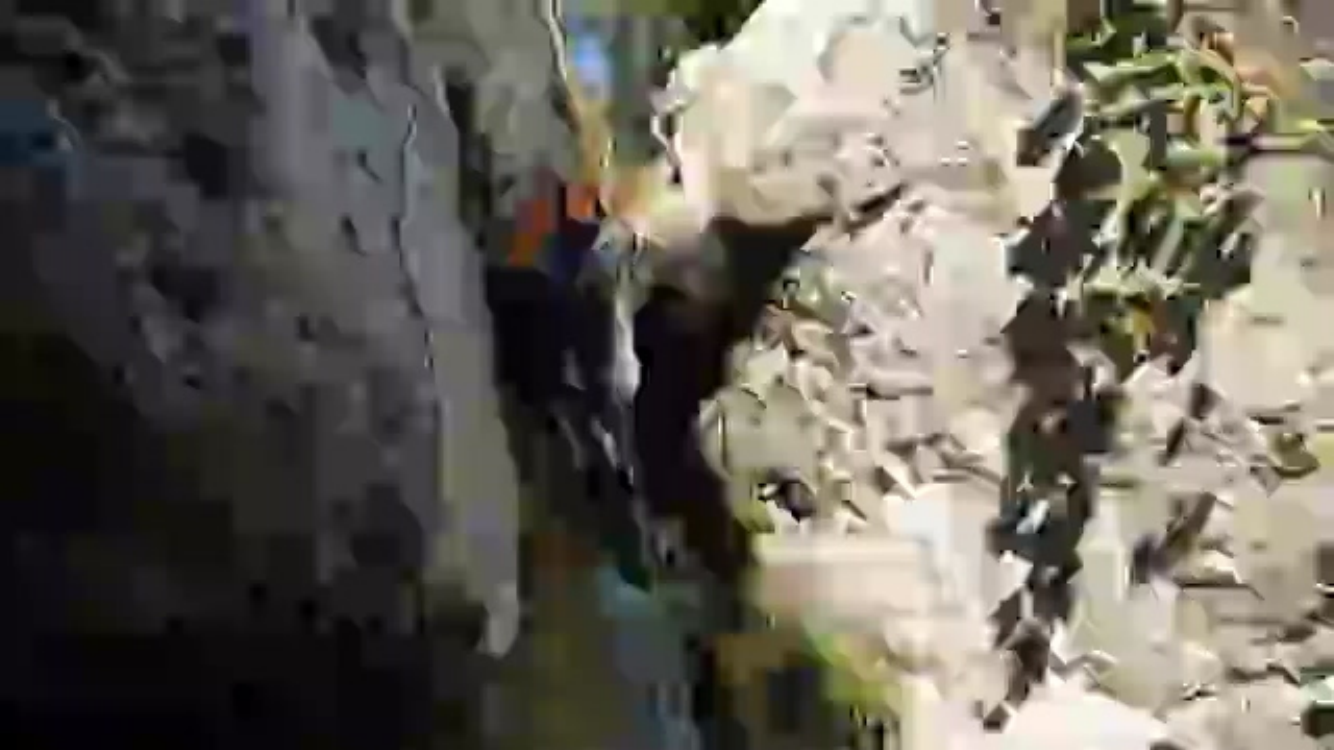}
            \label{fig:frame_gop1crf50}
        \end{minipage} \\
        
        \multicolumn{1}{c|}{2} &
        \begin{minipage}[t]{\myimgwidth\hsize}
            \centering
            \includegraphics[width=\linewidth]{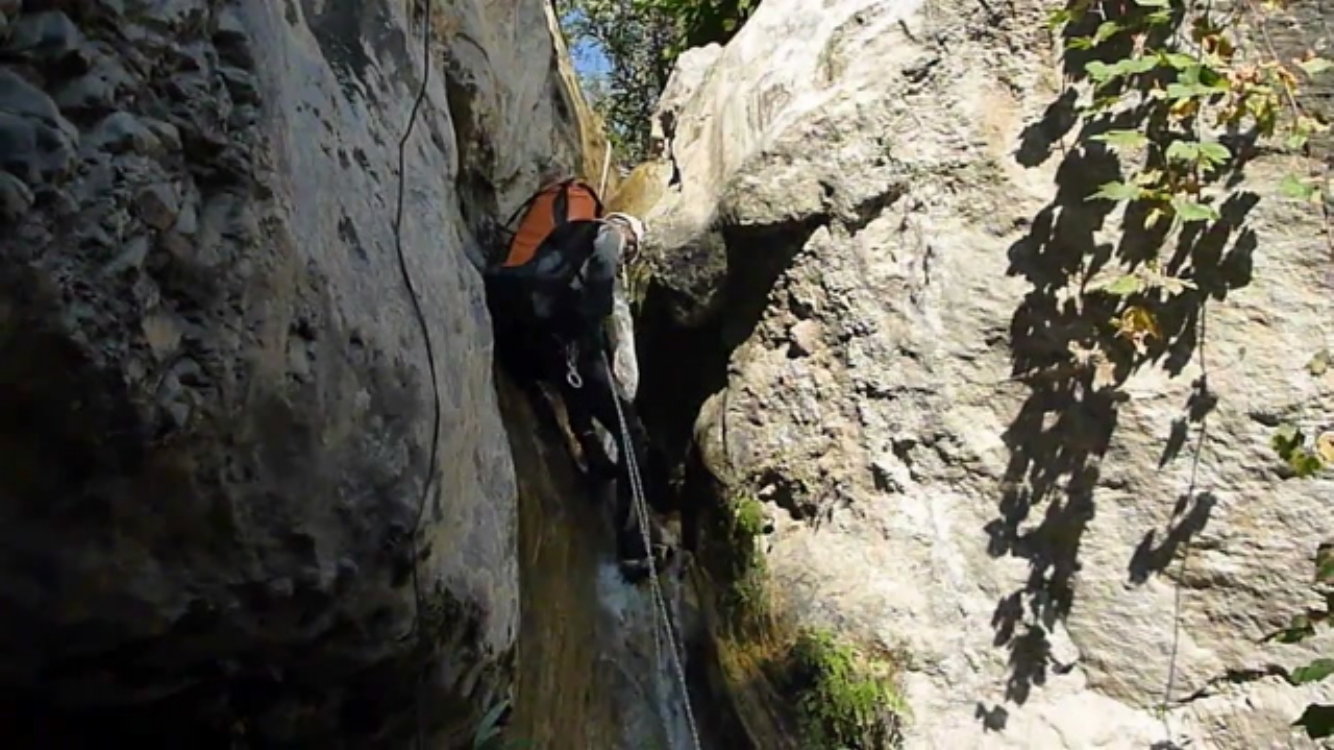}
            \label{fig:frame_gop2crf0}
        \end{minipage} &
        \begin{minipage}[t]{\myimgwidth\hsize}
            \centering
            \includegraphics[width=\linewidth]{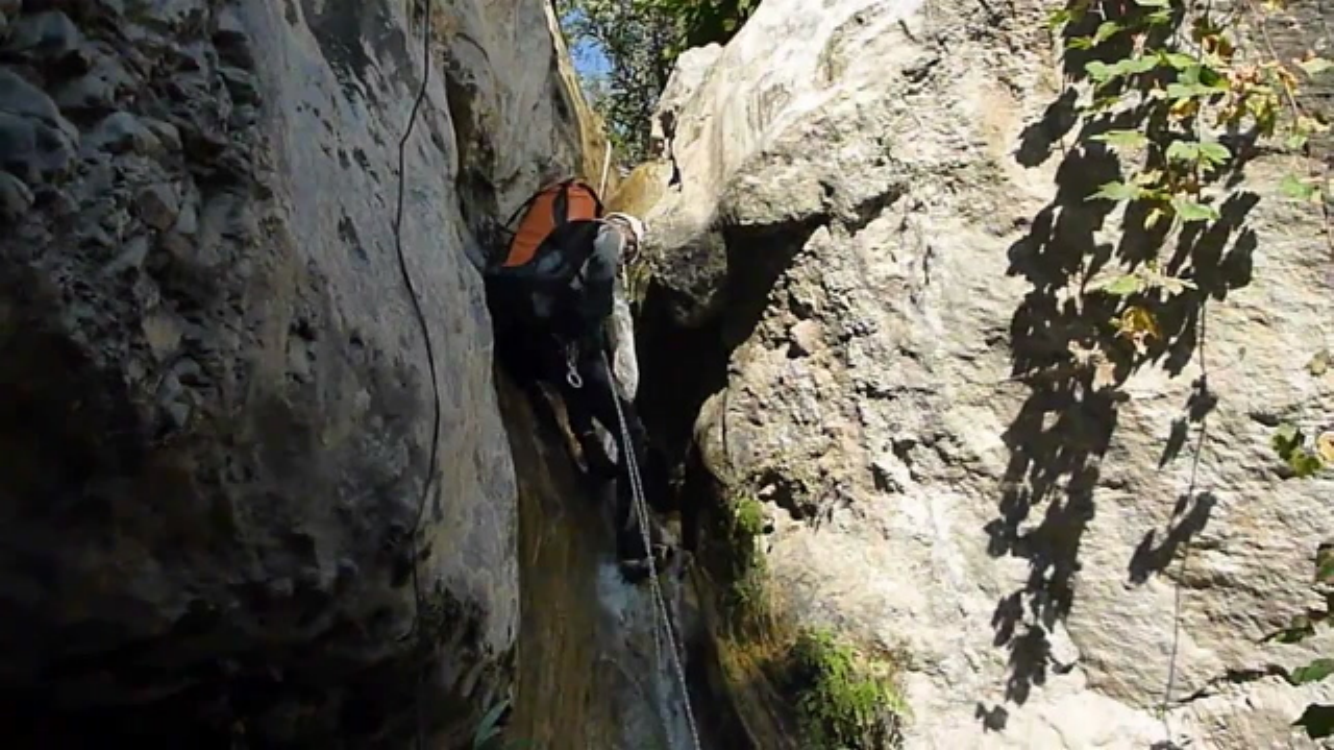}
            \label{fig:frame_gop2crf10}
        \end{minipage} &
        \begin{minipage}[t]{\myimgwidth\hsize}
            \centering
            \includegraphics[width=\linewidth]{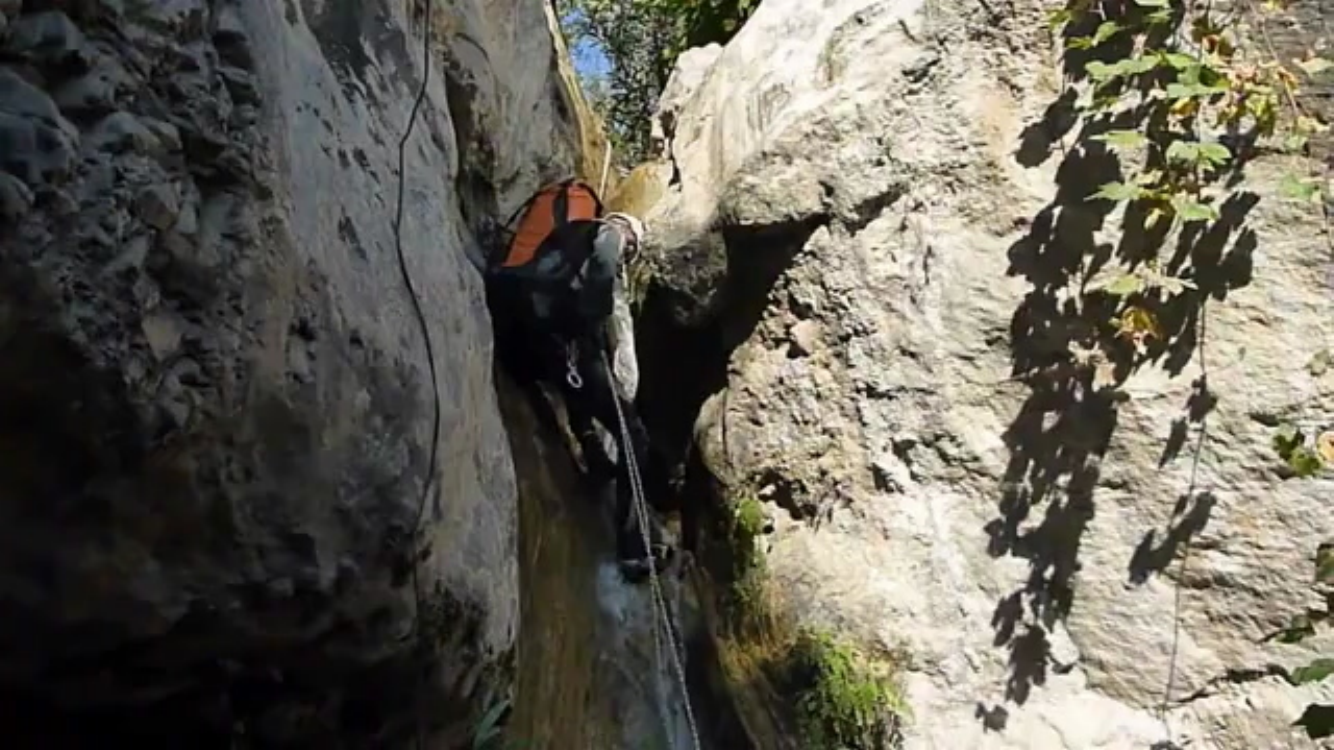}
            \label{fig:frame_gop2crf20}
        \end{minipage} &
        \begin{minipage}[t]{\myimgwidth\hsize}
            \centering
            \includegraphics[width=\linewidth]{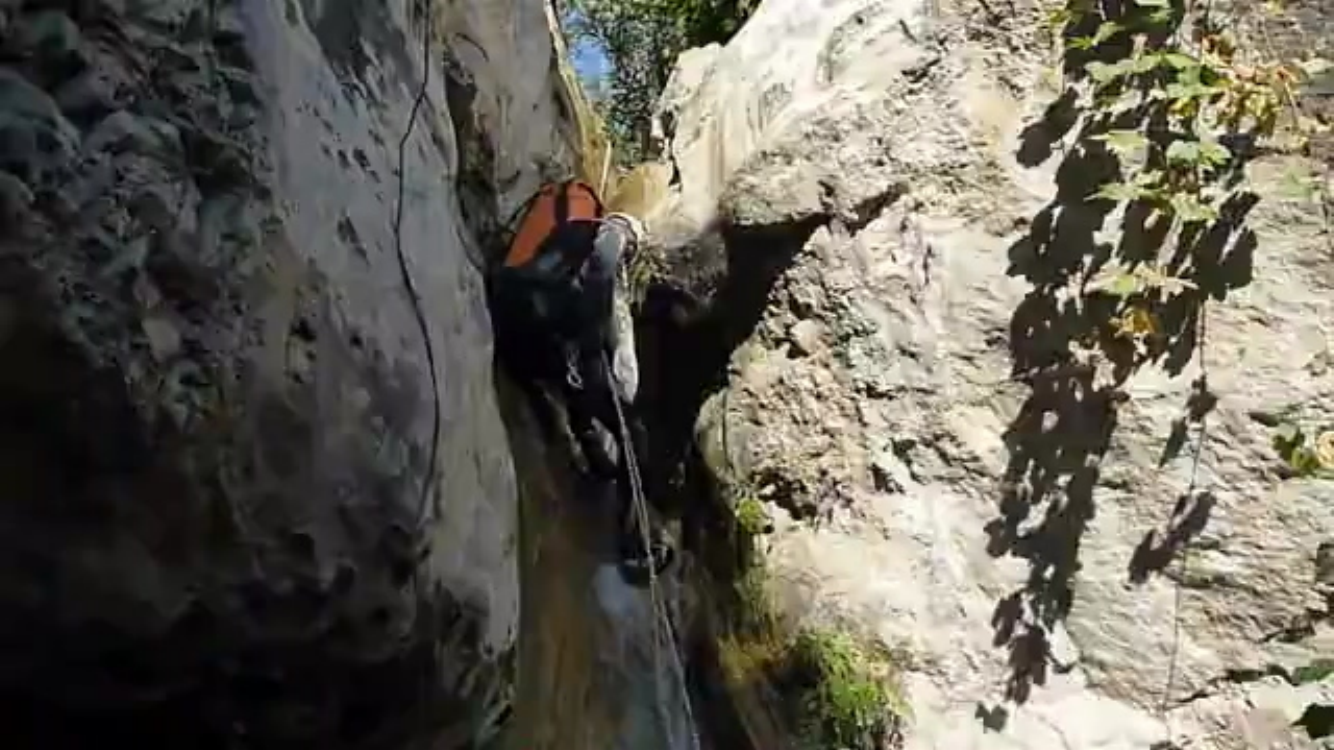}
            \label{fig:frame_gop2crf30}
        \end{minipage} &
        \begin{minipage}[t]{\myimgwidth\hsize}
            \centering
            \includegraphics[width=\linewidth]{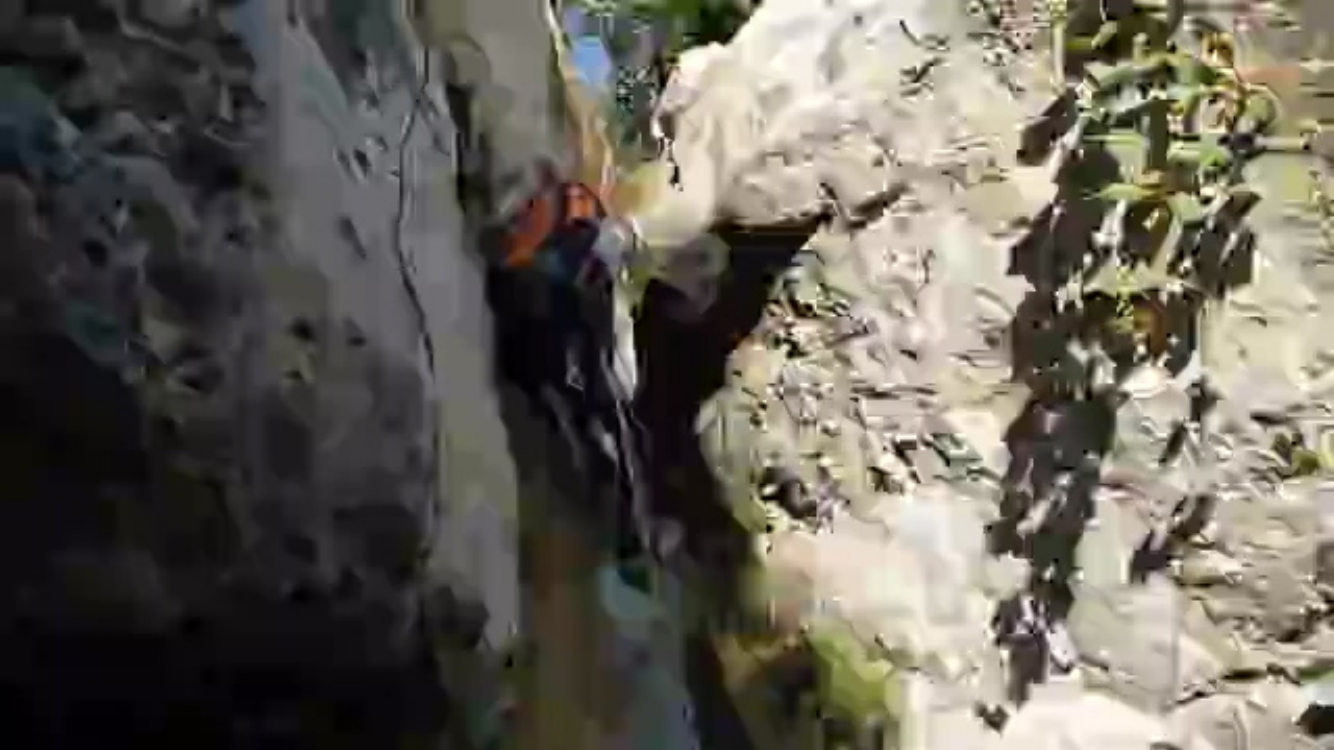}
            \label{fig:frame_gop2crf40}
        \end{minipage} &
        \begin{minipage}[t]{\myimgwidth\hsize}
            \centering
            \includegraphics[width=\linewidth]{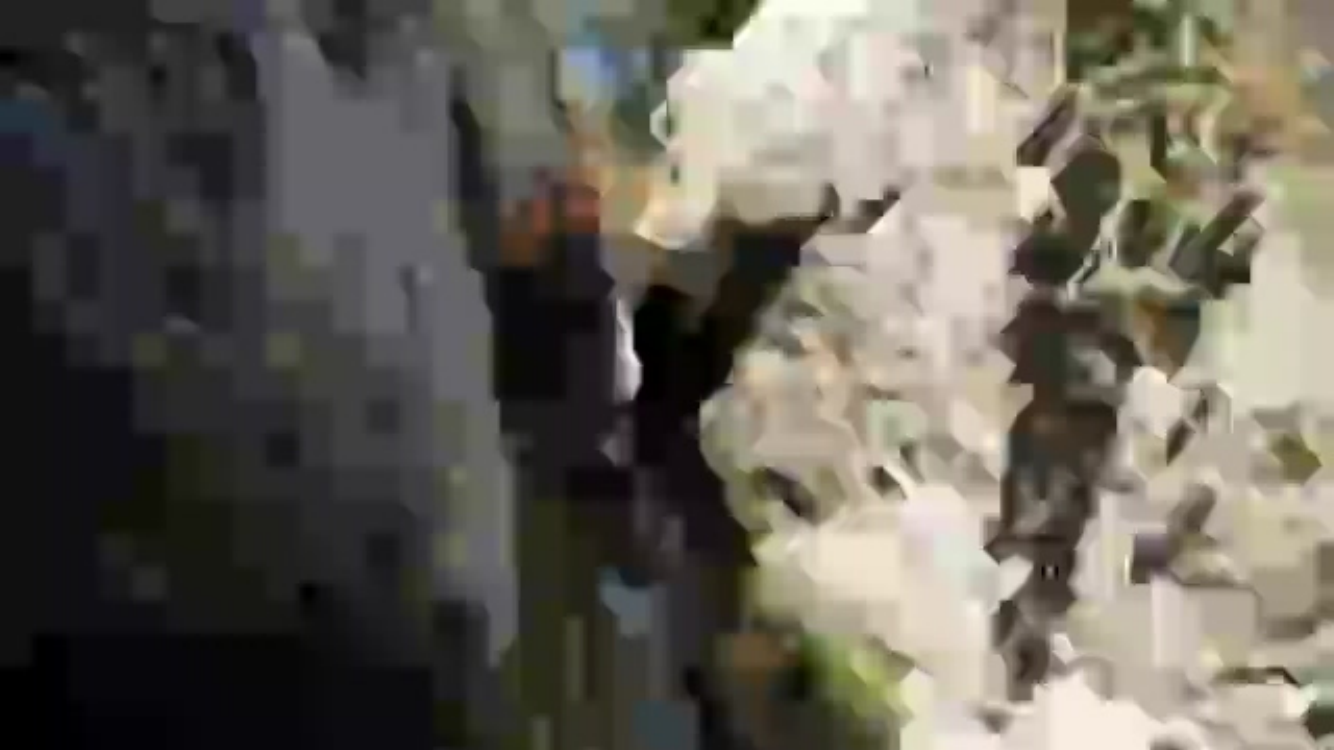}
            \label{fig:frame_gop2crf50}
        \end{minipage} \\
        
        \multicolumn{1}{c|}{5} &
        \begin{minipage}[t]{\myimgwidth\hsize}
            \centering
            \includegraphics[width=\linewidth]{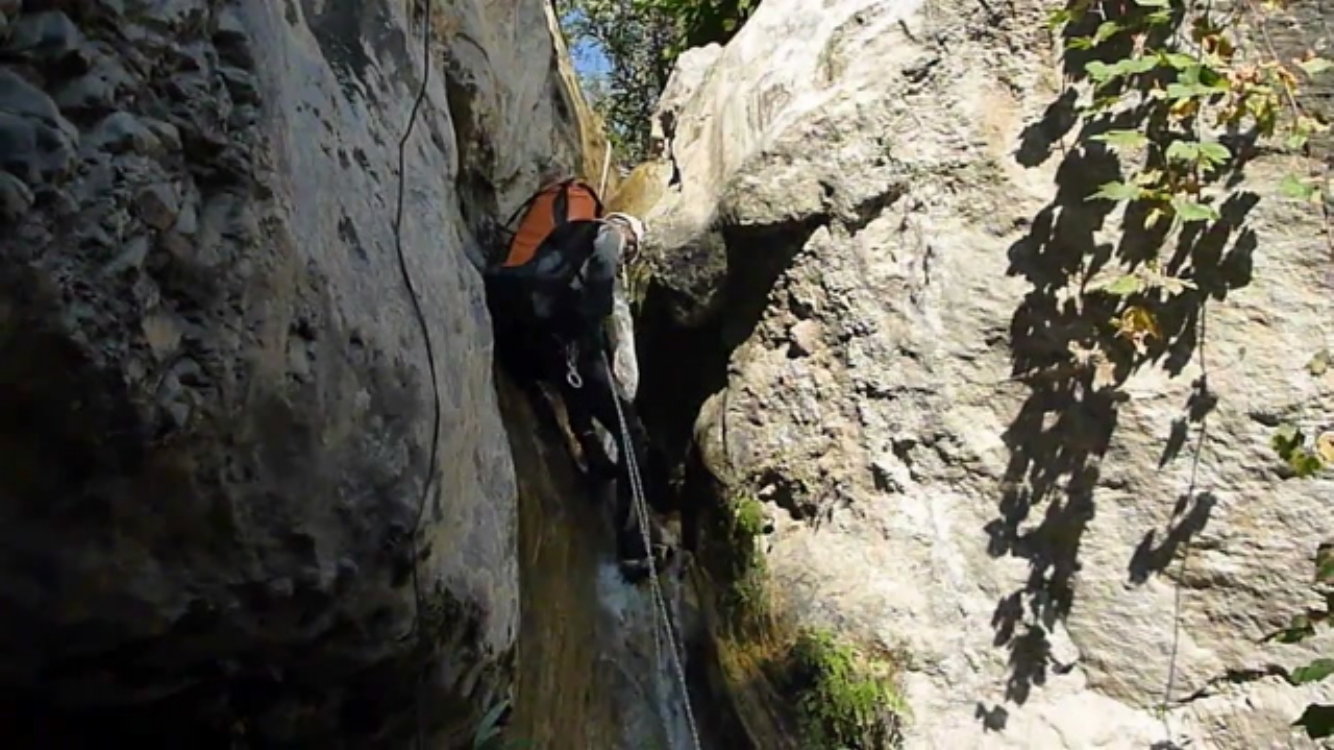}
            \label{fig:frame_gop5crf0}
        \end{minipage} &
        \begin{minipage}[t]{\myimgwidth\hsize}
            \centering
            \includegraphics[width=\linewidth]{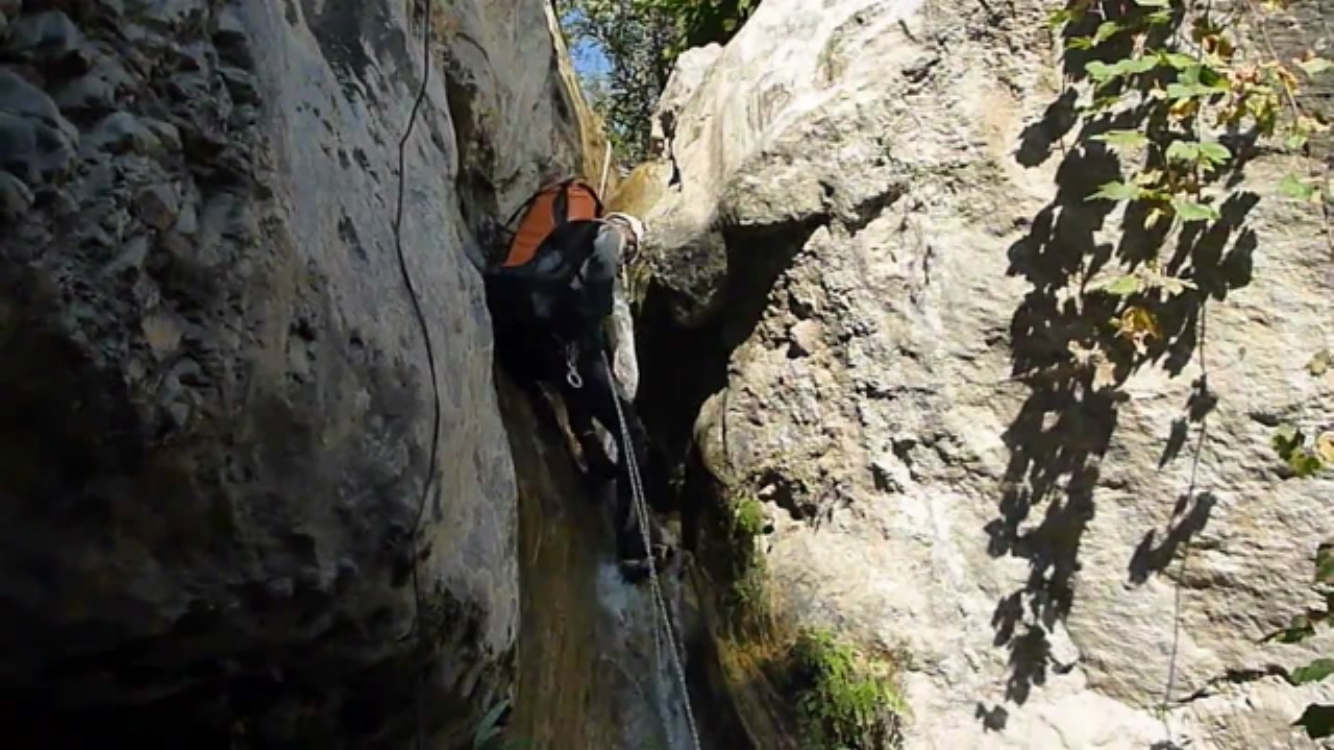}
            \label{fig:frame_gop5crf10}
        \end{minipage} &
        \begin{minipage}[t]{\myimgwidth\hsize}
            \centering
            \includegraphics[width=\linewidth]{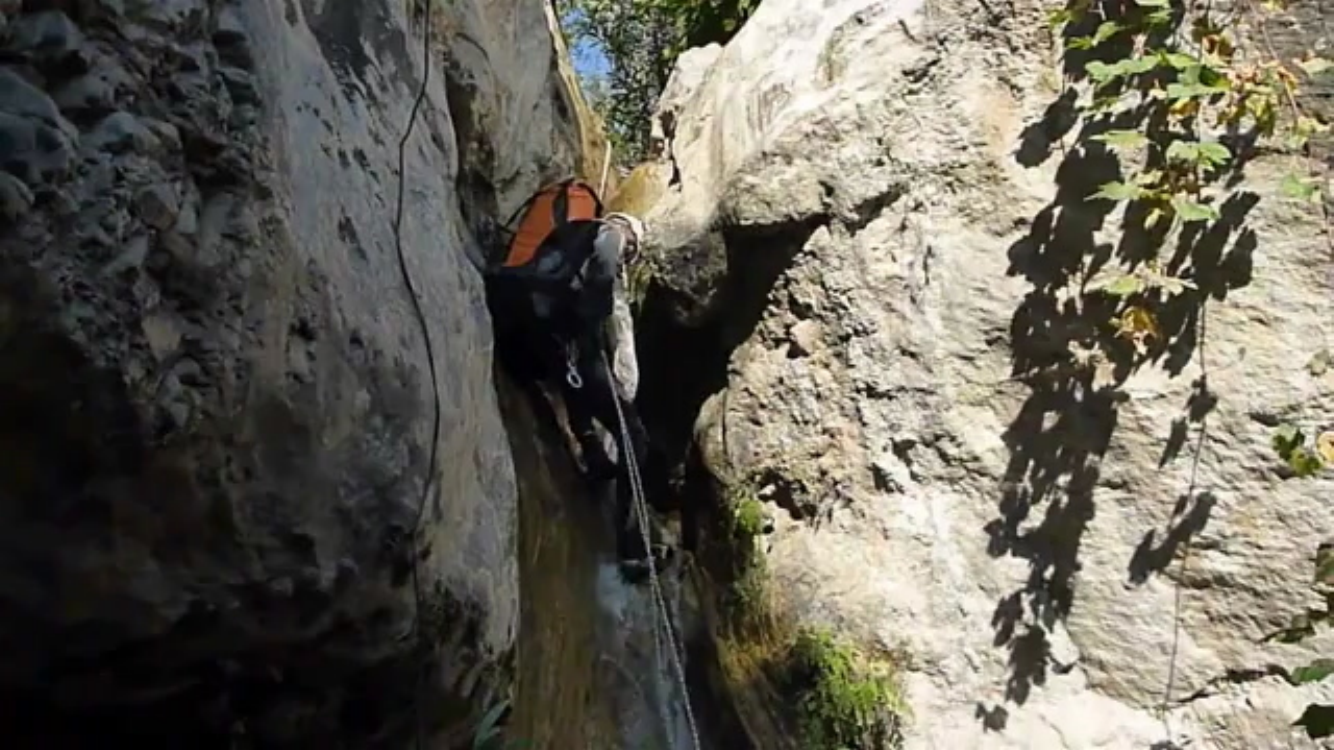}
            \label{fig:frame_gop5crf20}
        \end{minipage} &
        \begin{minipage}[t]{\myimgwidth\hsize}
            \centering
            \includegraphics[width=\linewidth]{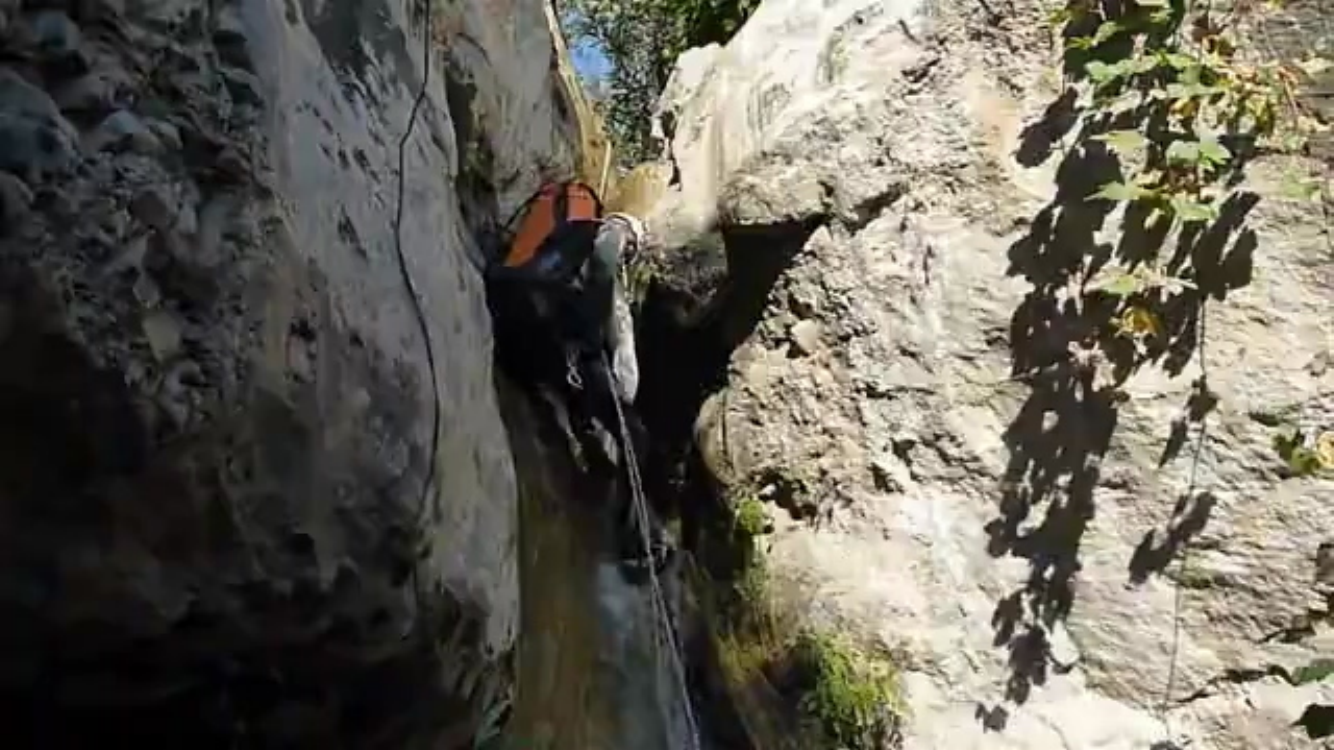}
            \label{fig:frame_gop5crf30}
        \end{minipage} &
        \begin{minipage}[t]{\myimgwidth\hsize}
            \centering
            \includegraphics[width=\linewidth]{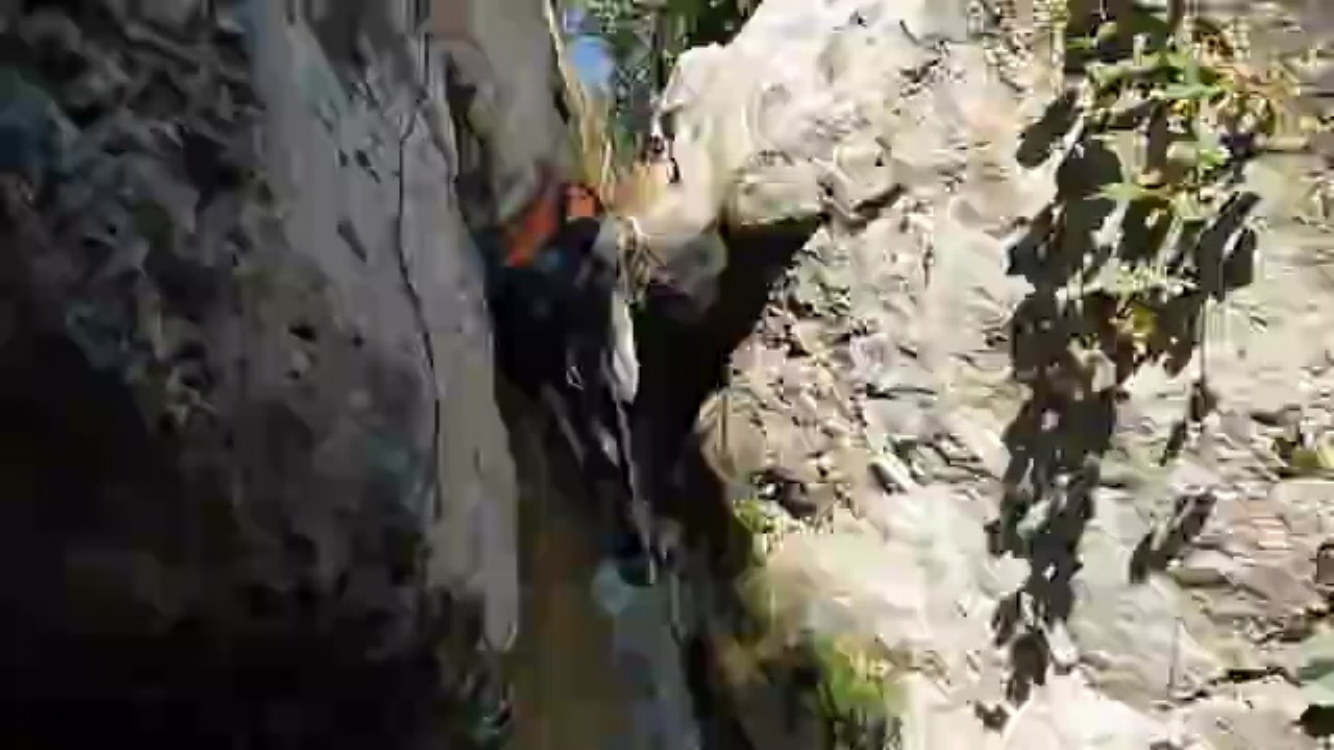}
            \label{fig:frame_gop5crf40}
        \end{minipage} &
        \begin{minipage}[t]{\myimgwidth\hsize}
            \centering
            \includegraphics[width=\linewidth]{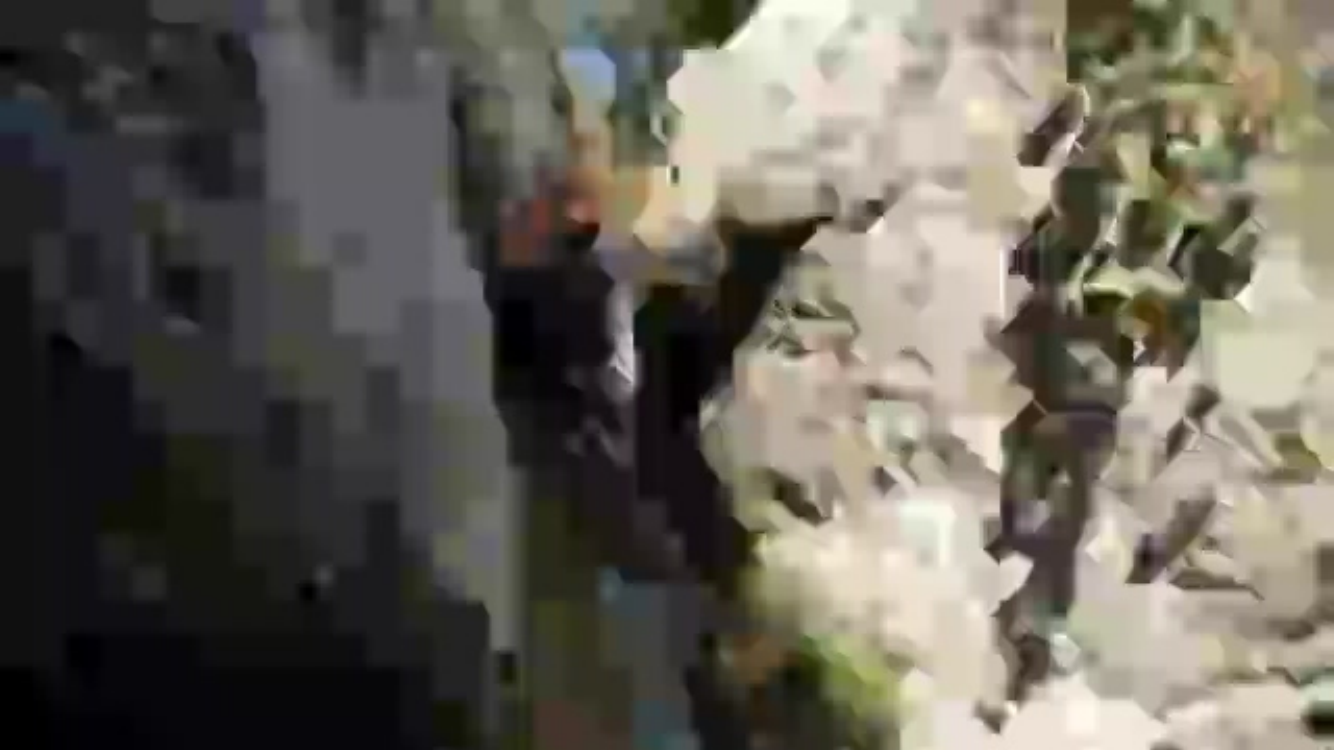}
            \label{fig:frame_gop5crf50}
        \end{minipage} \\
        
        \multicolumn{1}{c|}{10} &
        \begin{minipage}[t]{\myimgwidth\hsize}
            \centering
            \includegraphics[width=\linewidth]{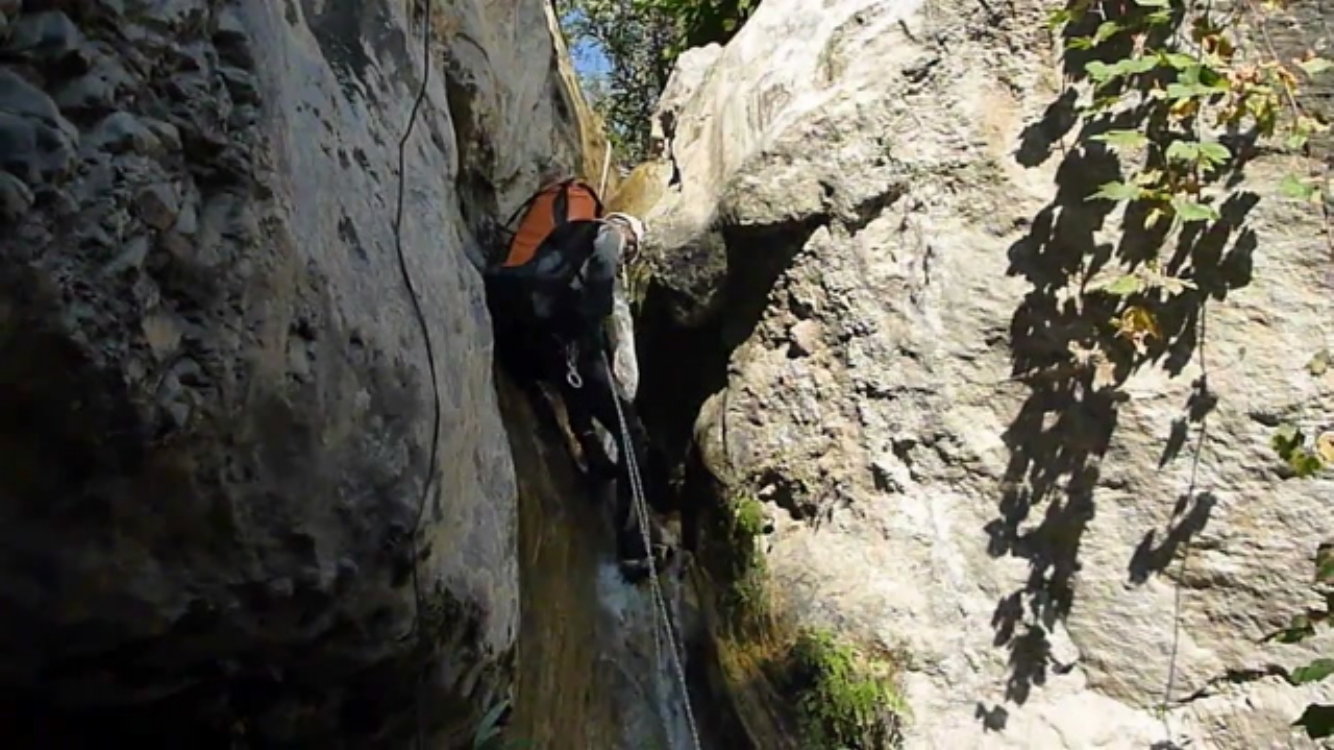}
            \label{fig:frame_gop10crf0}
        \end{minipage} &
        \begin{minipage}[t]{\myimgwidth\hsize}
            \centering
            \includegraphics[width=\linewidth]{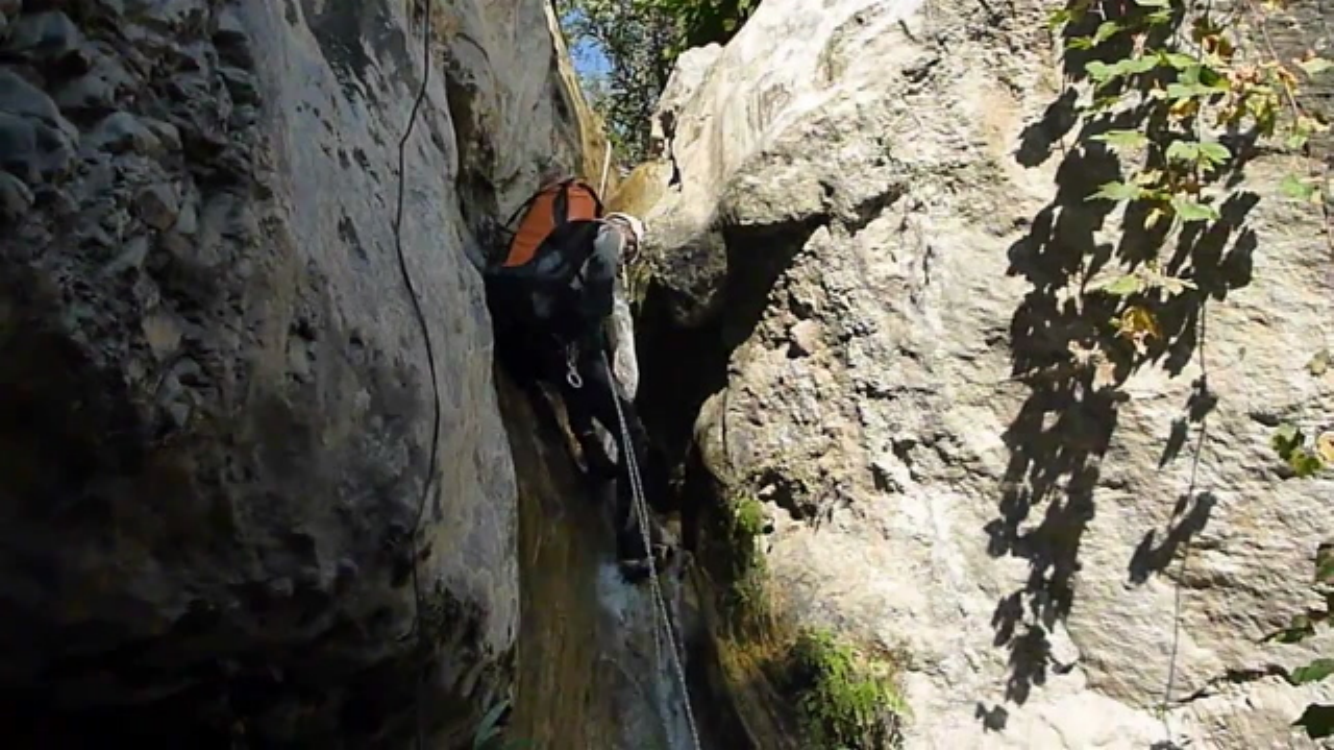}
            \label{fig:frame_gop10crf10}
        \end{minipage} &
        \begin{minipage}[t]{\myimgwidth\hsize}
            \centering
            \includegraphics[width=\linewidth]{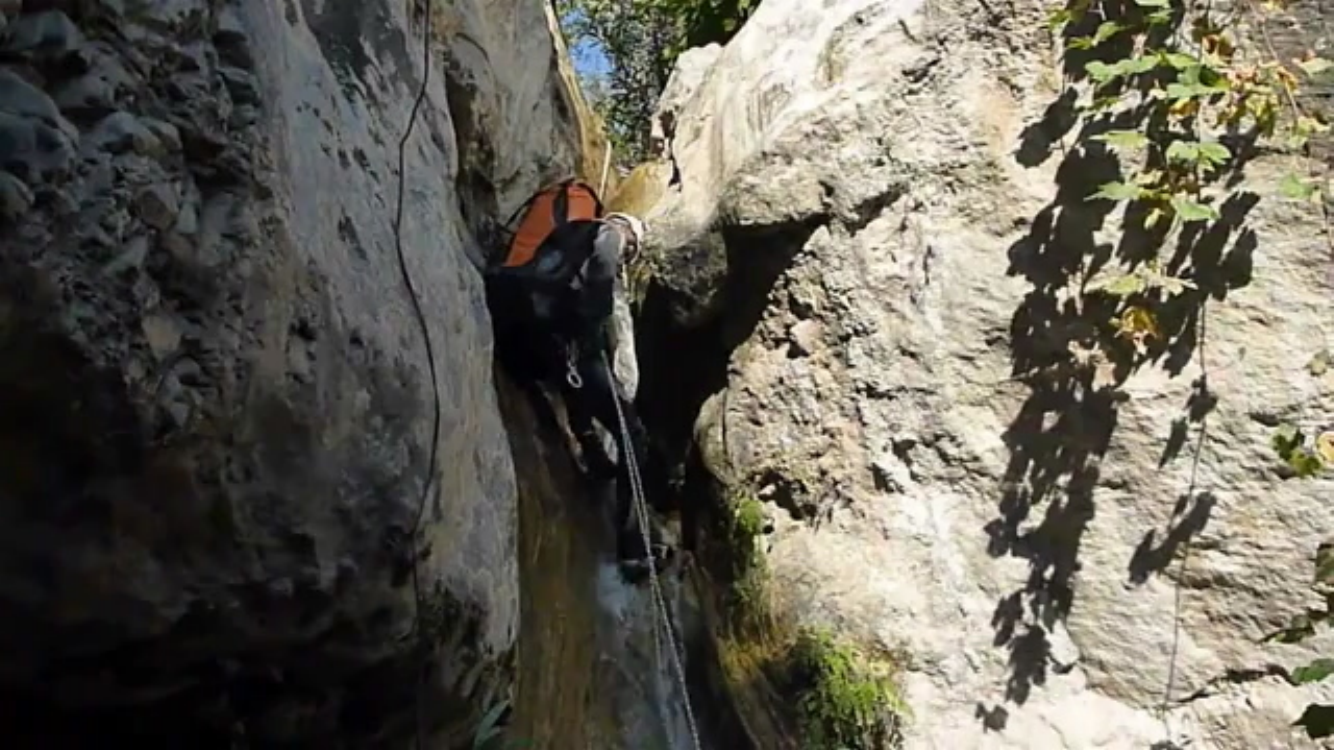}
            \label{fig:frame_gop10crf20}
        \end{minipage} &
        \begin{minipage}[t]{\myimgwidth\hsize}
            \centering
            \includegraphics[width=\linewidth]{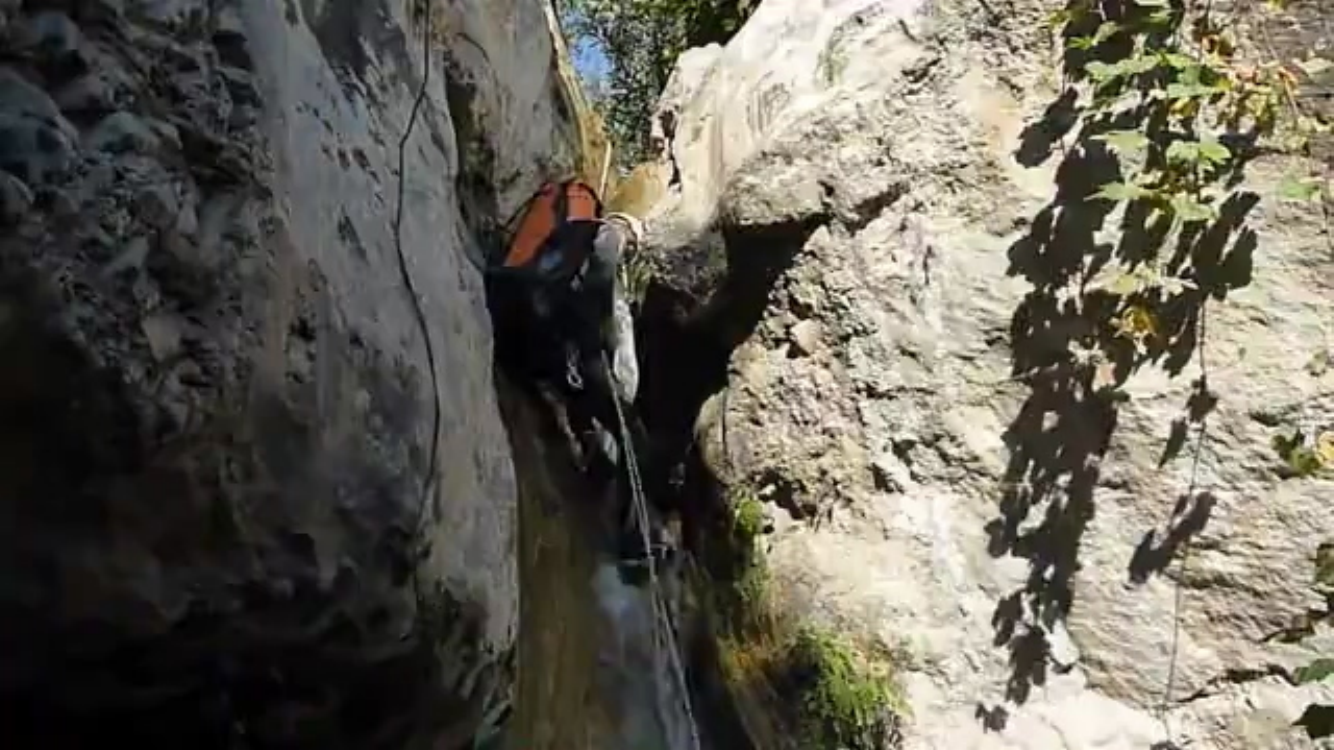}
            \label{fig:frame_gop10crf30}
        \end{minipage} &
        \begin{minipage}[t]{\myimgwidth\hsize}
            \centering
            \includegraphics[width=\linewidth]{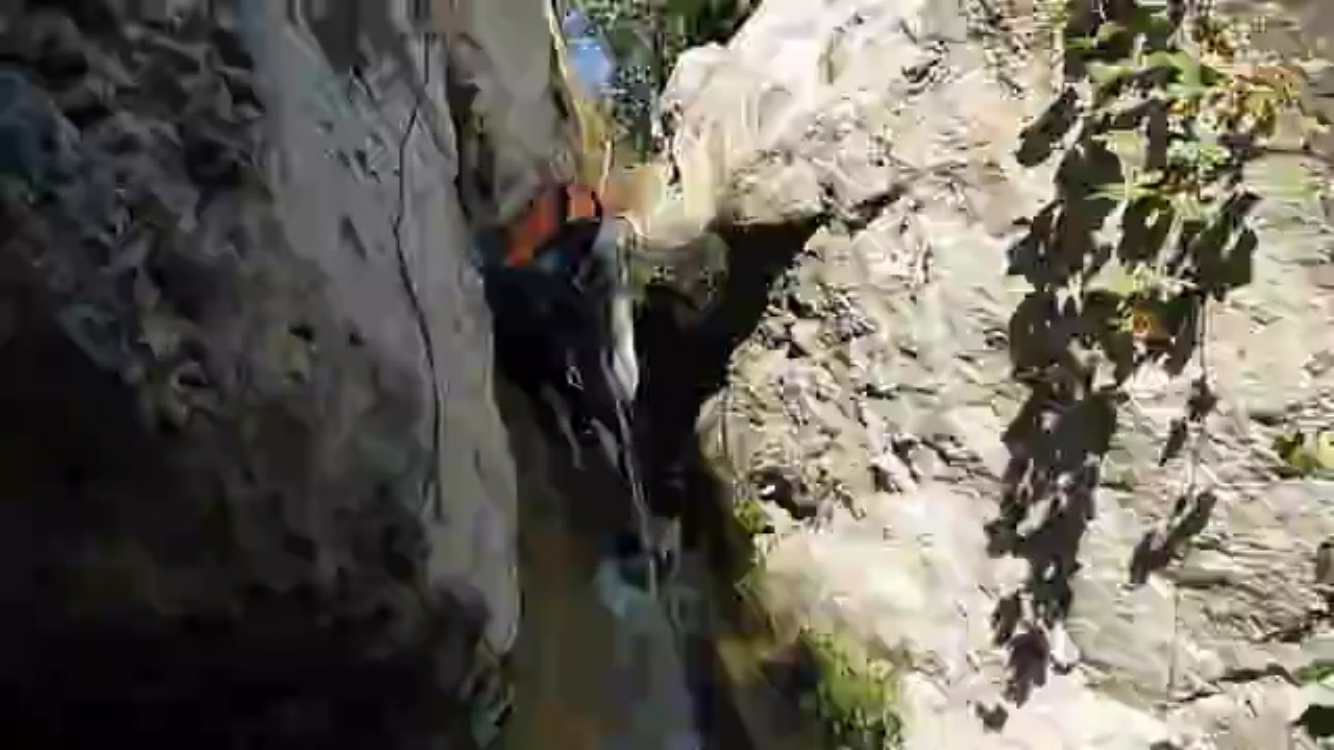}
            \label{fig:frame_gop10crf40}
        \end{minipage} &
        \begin{minipage}[t]{\myimgwidth\hsize}
            \centering
            \includegraphics[width=\linewidth]{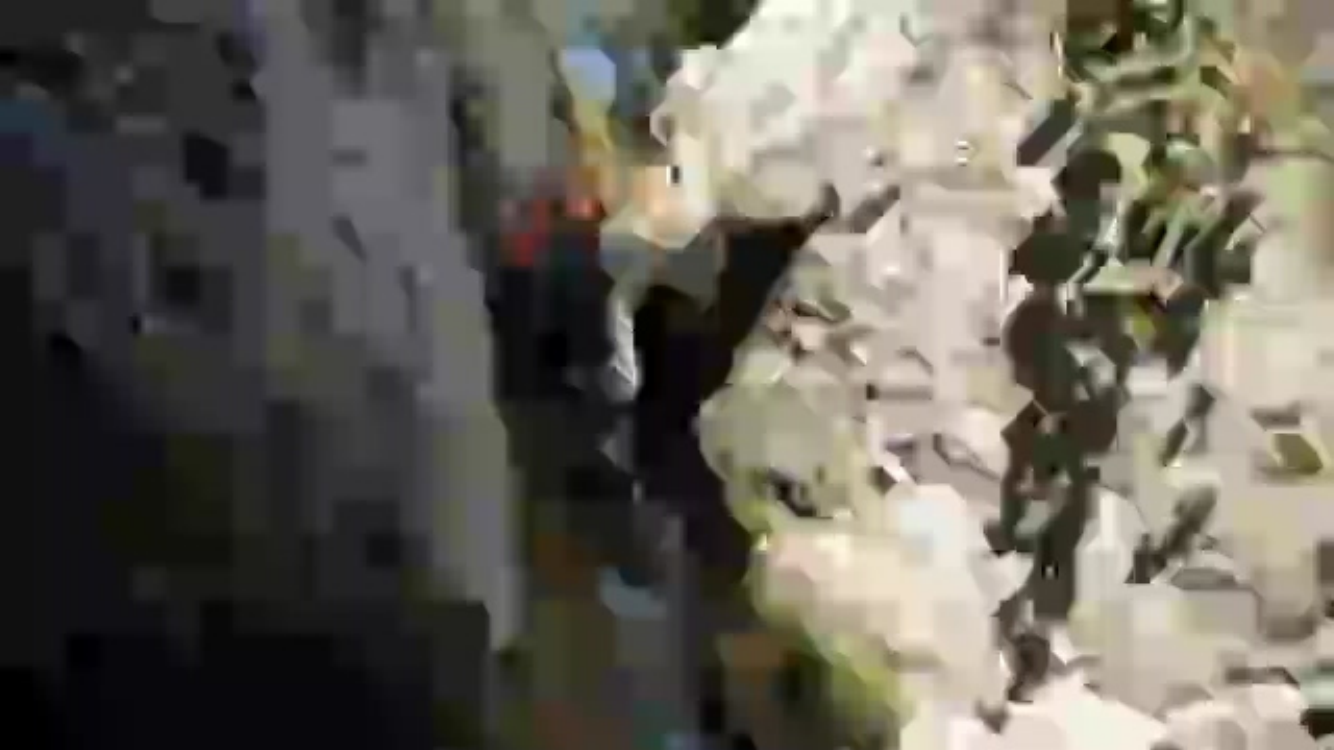}
            \label{fig:frame_gop10crf50}
        \end{minipage} \\
        
        \multicolumn{1}{c|}{12} &
        \begin{minipage}[t]{\myimgwidth\hsize}
            \centering
            \includegraphics[width=\linewidth]{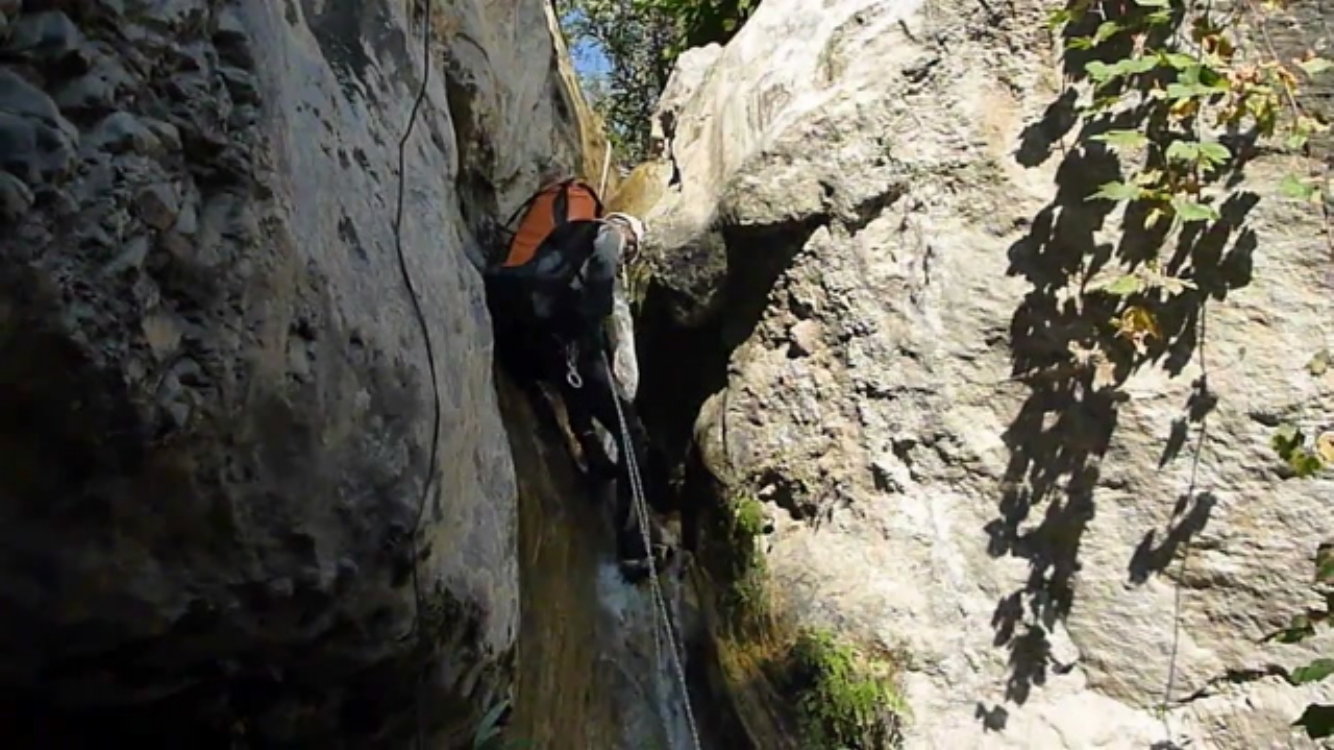}
            \label{fig:frame_gop12crf0}
        \end{minipage} &
        \begin{minipage}[t]{\myimgwidth\hsize}
            \centering
            \includegraphics[width=\linewidth]{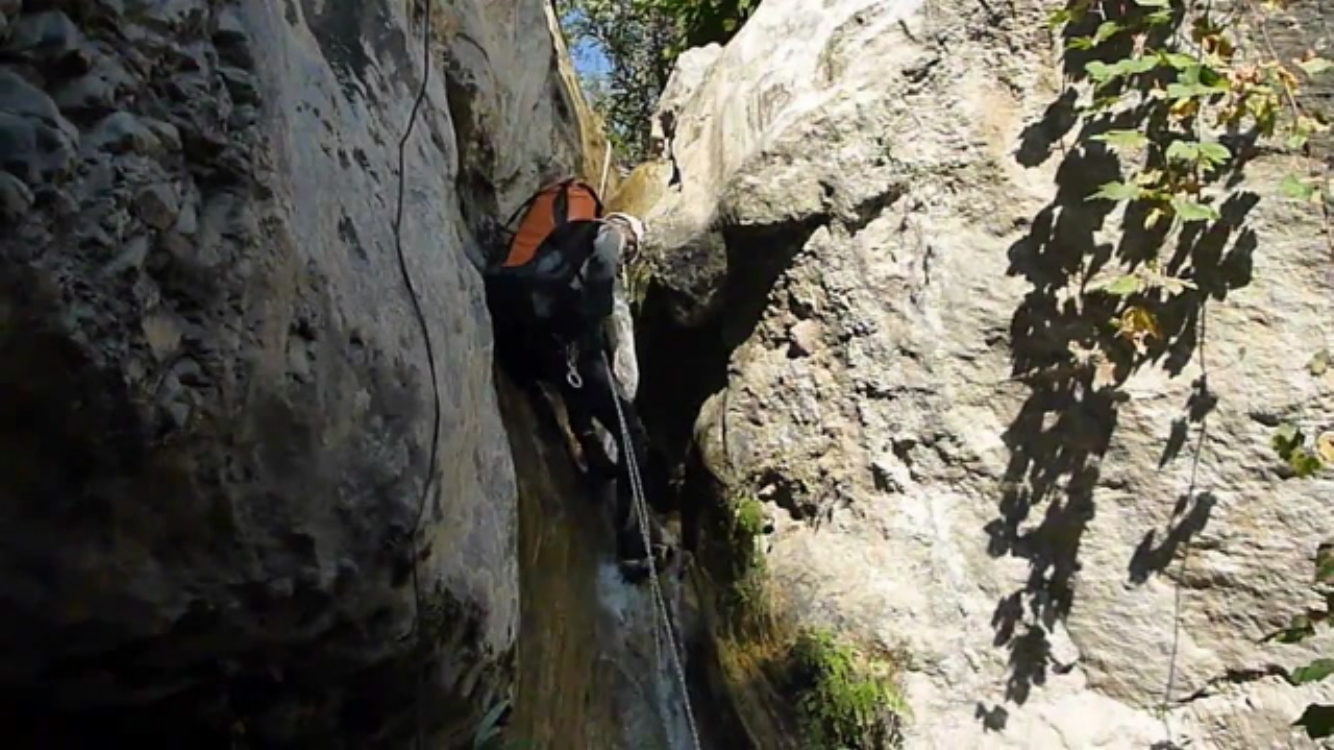}
            \label{fig:frame_gop12crf10}
        \end{minipage} &
        \begin{minipage}[t]{\myimgwidth\hsize}
            \centering
            \includegraphics[width=\linewidth]{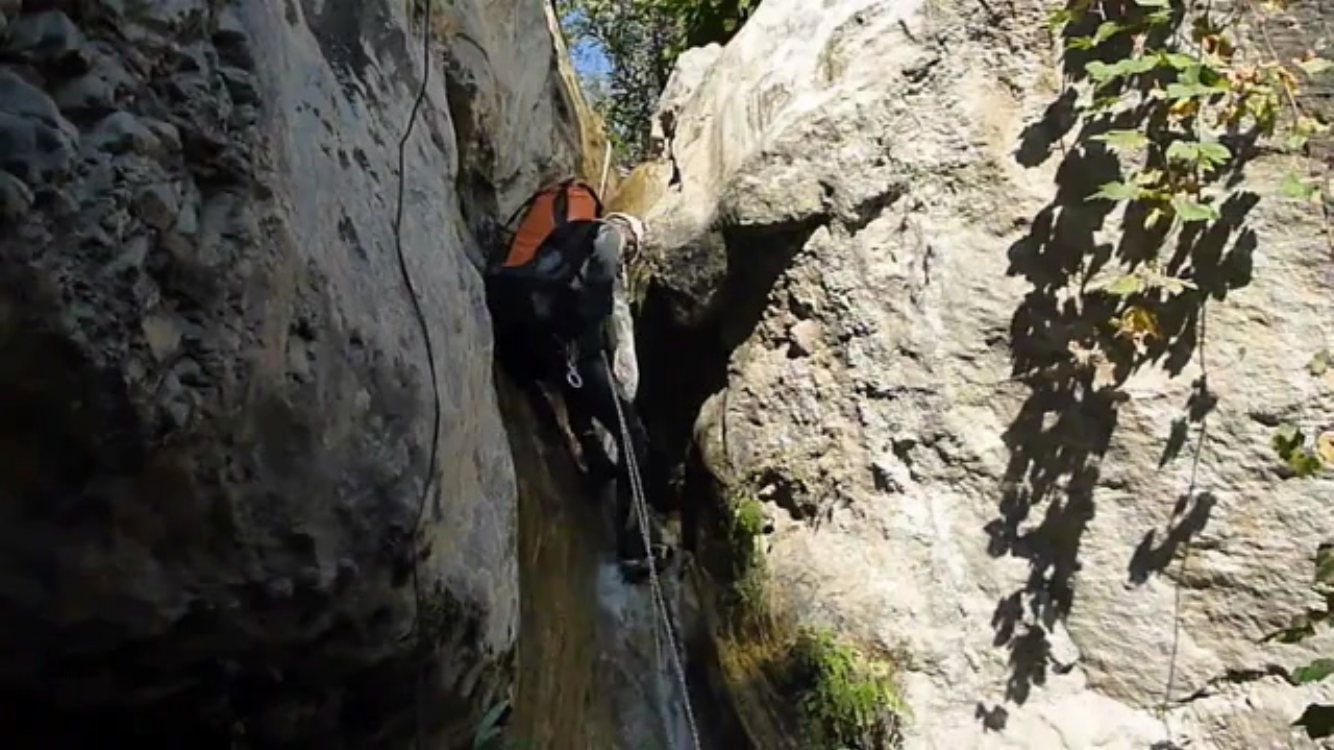}
            \label{fig:frame_gop12crf20}
        \end{minipage} &
        \begin{minipage}[t]{\myimgwidth\hsize}
            \centering
            \includegraphics[width=\linewidth]{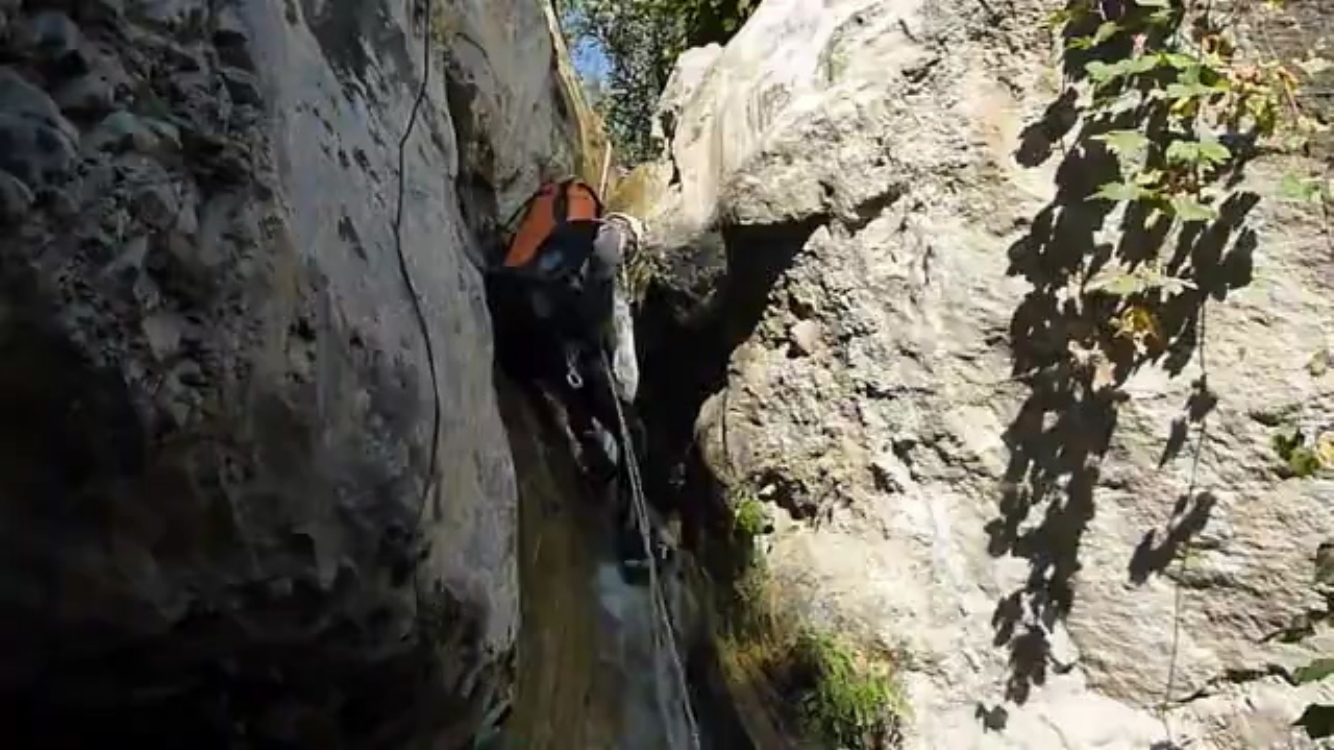}
            \label{fig:frame_gop12crf30}
        \end{minipage} &
        \begin{minipage}[t]{\myimgwidth\hsize}
            \centering
            \includegraphics[width=\linewidth]{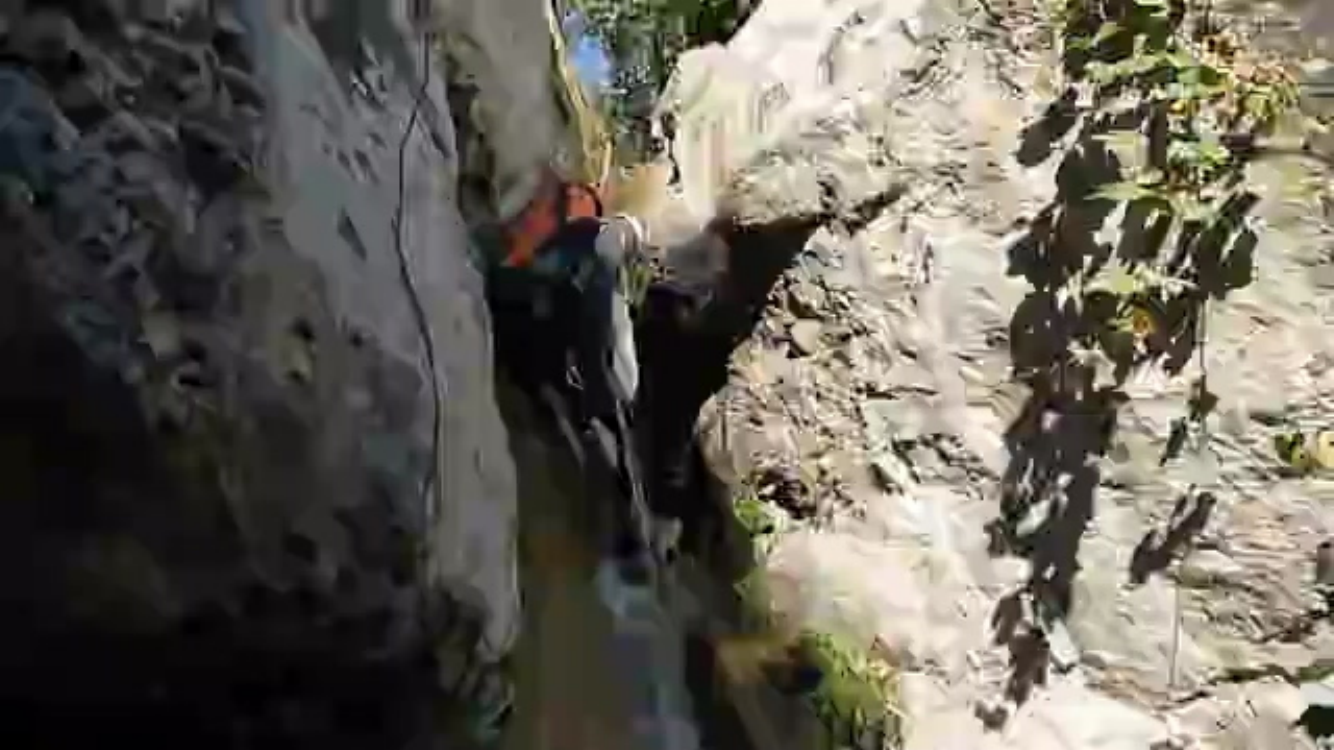}
            \label{fig:frame_gop12crf40}
        \end{minipage} &
        \begin{minipage}[t]{\myimgwidth\hsize}
            \centering
            \includegraphics[width=\linewidth]{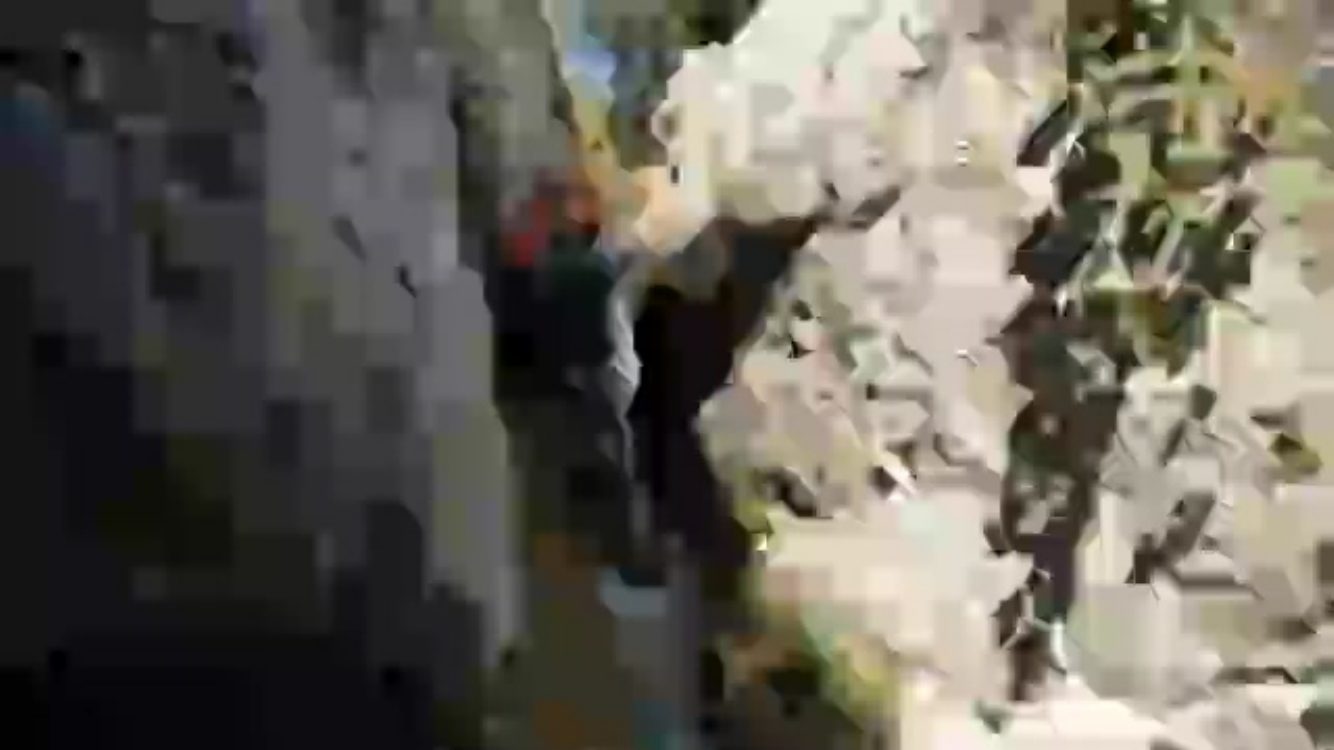}
            \label{fig:frame_gop12crf50}
        \end{minipage} \\
        
    \end{tabular}
    
    }

    \caption{Changes of image quality due to H.264/AVC transcoding
    with different GOP sizes and CRF values.
    }
    \label{fig:ffmpeg_frame}

\end{figure*}

%% file: fig/acc_jpeg_compression.tex
\begin{figure}[t]
    \centering
    \def\myimgwidth{0.48}
    \def\myimgsclae{0.26}

    \begin{tabular}{@{}c@{}c@{}}
        \begin{minipage}[t]{\myimgwidth\hsize}
            \centering
            \includegraphics[scale=\myimgsclae]{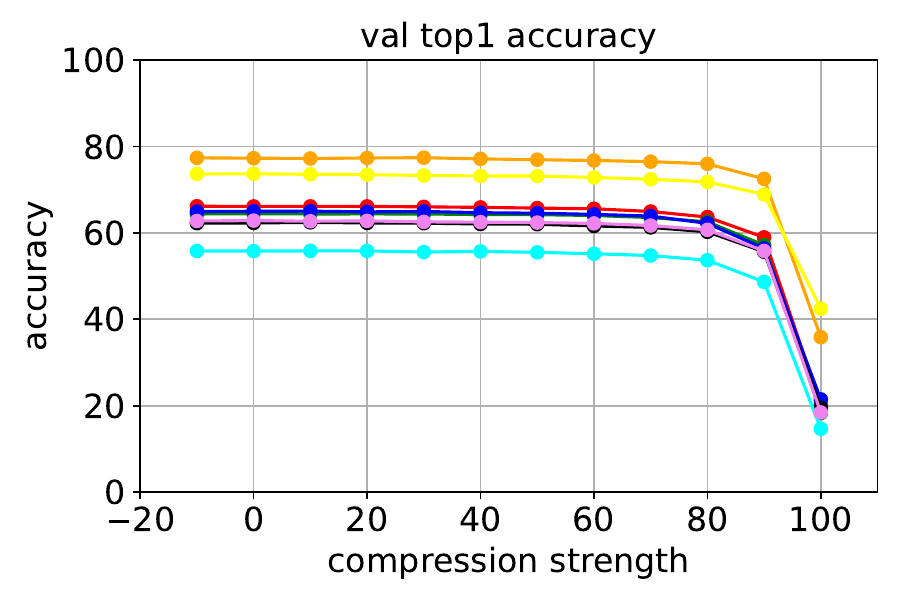}
            \begin{overpic}[scale=\myimgsclae,percent]{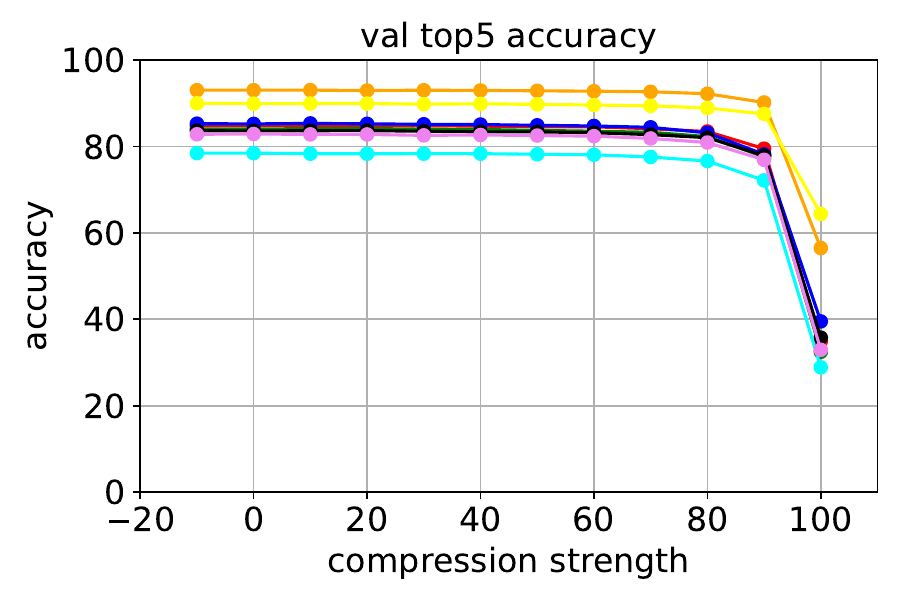}
                \put(20,15){
                    \includegraphics[width=0.2\linewidth]{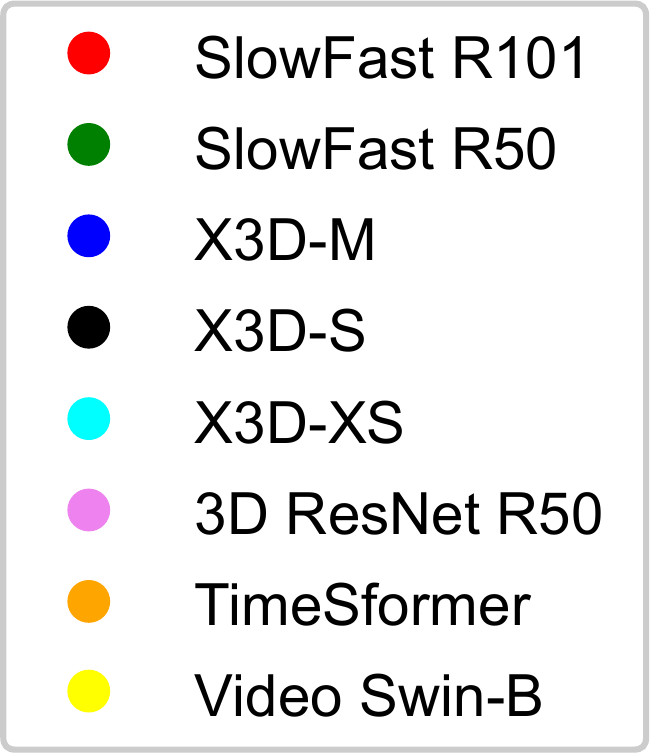}
                }
            \end{overpic}
            \subcaption{}
            \label{fig:acc_jpeg_compression_-10_100}
            \label{fig:top5_-10_100}            
            \label{fig:top1_-10_100}
        \end{minipage}
        &
        \begin{minipage}[t]{\myimgwidth\hsize}
            \centering
            \includegraphics[scale=\myimgsclae]{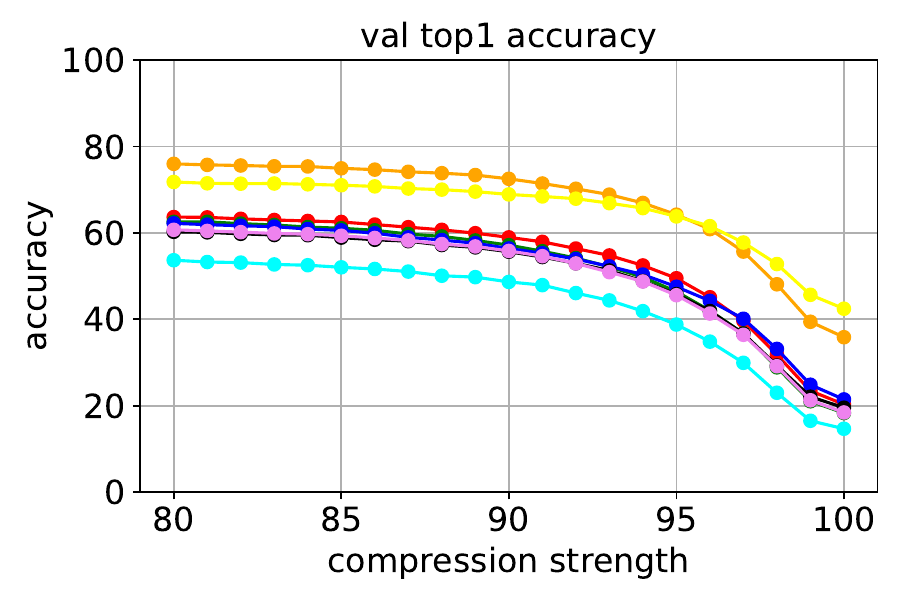}
            \includegraphics[scale=\myimgsclae]{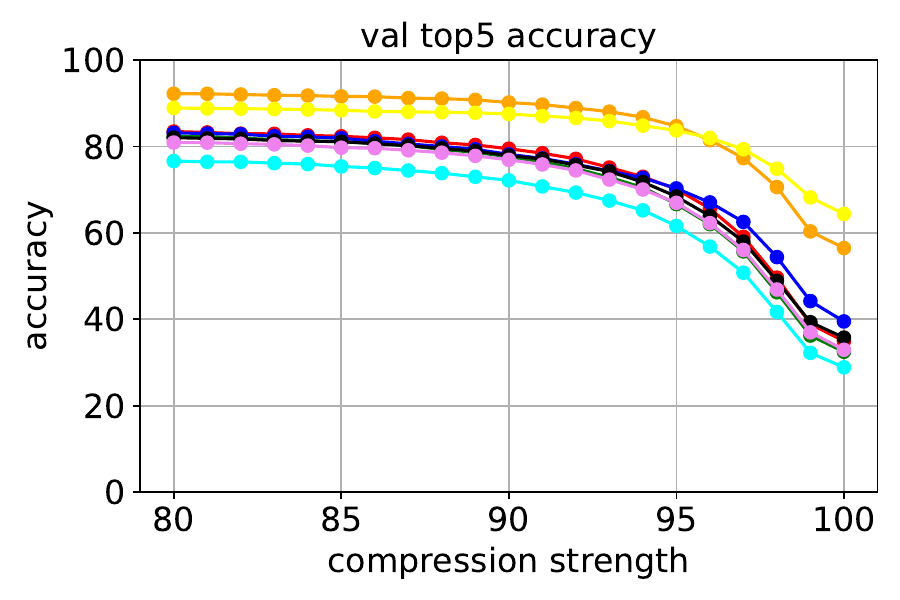}
            \subcaption{}
            \label{fig:acc_jpeg_compression_80_100}
            \label{fig:top1_80_100}
            \label{fig:top5_80_100}
        \end{minipage} 
    \end{tabular}


    \caption{Top-1 (top) and top-5 (bottom) performance of each model for videos degraded by JPEG transcoding.
    (a) Compression strength $\cs$ increased by 10 from 0 to 100, and
    (b) increased by 1 from 80 to 100.
    Note that $\cs=-10$ represents the original video.
    }
    \label{fig:acc_jpeg_compression}

\end{figure}

%% file: fig/acc_jpeg_image_score.tex
\begin{figure}[t]
    \centering
    \def\myimgwidth{0.48}
    \def\myimgsclae{0.27}
    
    \begin{tabular}{@{}c@{}c@{}}
        \begin{minipage}[t]{\myimgwidth\hsize}
            \centering
            \begin{overpic}[scale=0.2415,percent]{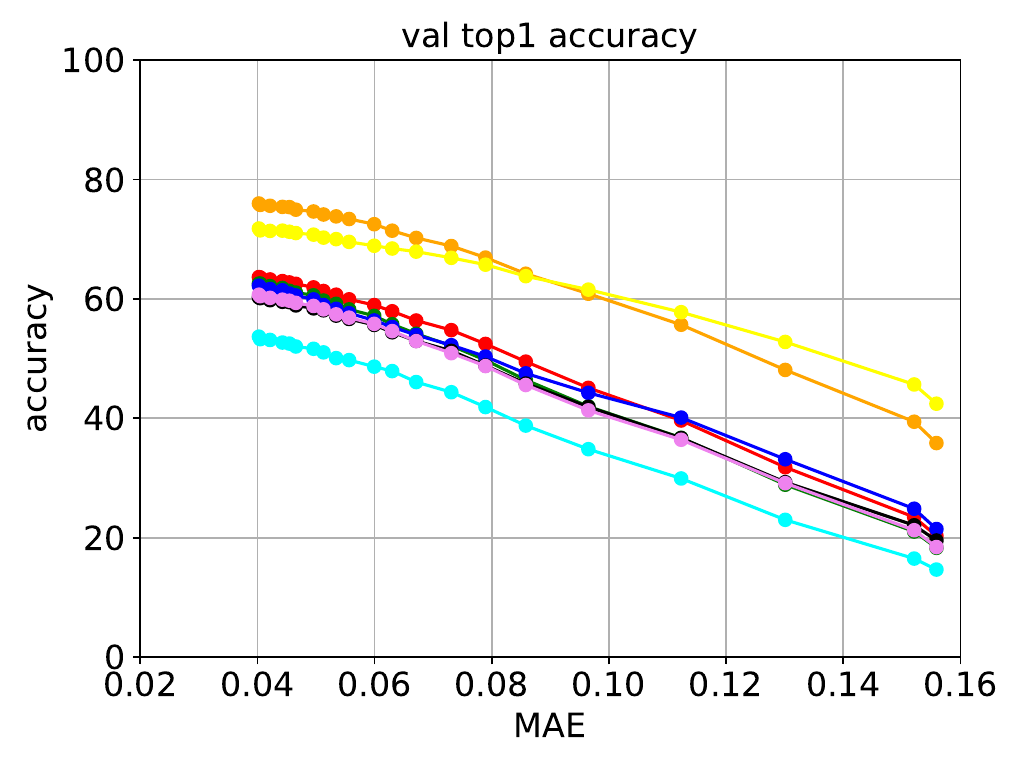}
                    \put(15,12){
                    \includegraphics[width=0.2\linewidth]{images/acc_jpeg/image_score/80_100/legend.pdf}
                }
            \end{overpic}
            \includegraphics[width=\linewidth]{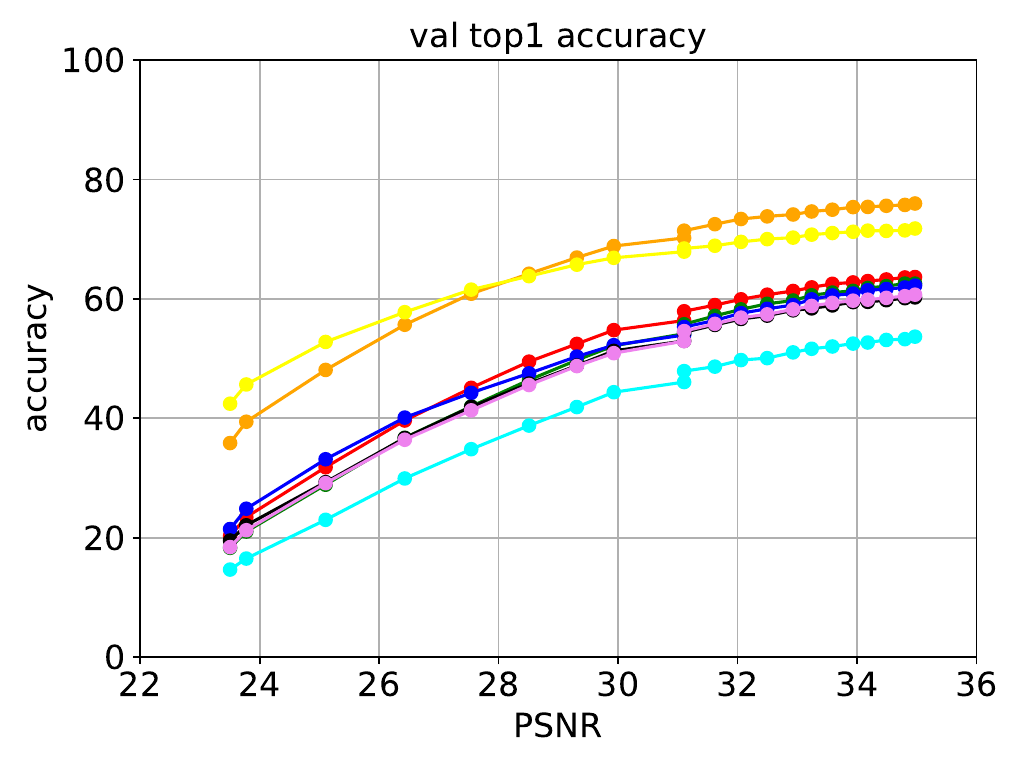}\\
        \end{minipage}
         &
        \begin{minipage}[t]{\myimgwidth\hsize}
            \centering
            \includegraphics[width=\linewidth]{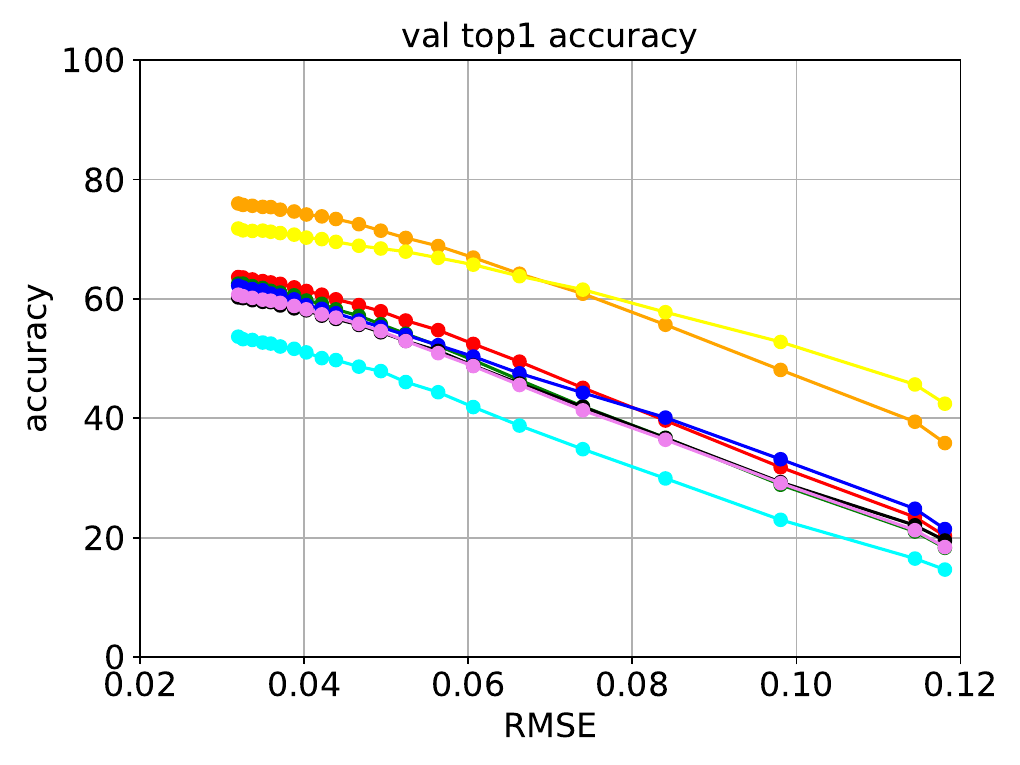}\\
            \includegraphics[width=\linewidth]{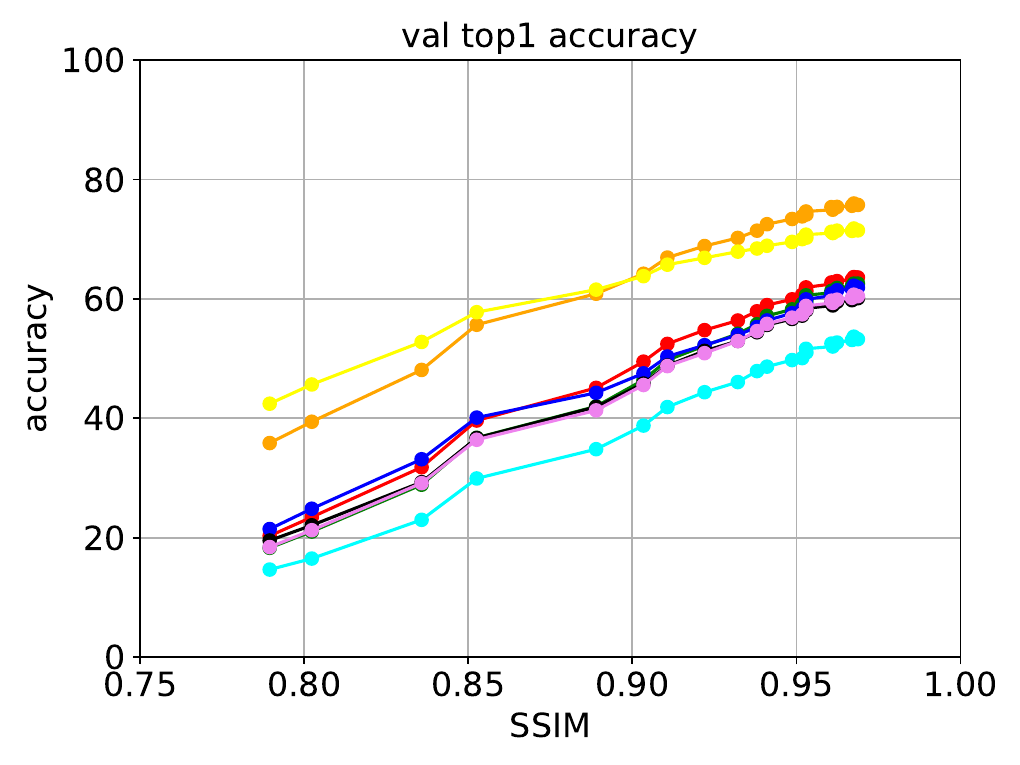}
        \end{minipage}

    \end{tabular}

    \caption{Relationship between top-1 performance 
    and quality
    (MAE, RMSE, PSNR, and SSIM) of JPEG transcoding
    ($\cs$ from 80 to 100).
    For MAE and RMSE, smaller values indicate higher quality,
    while larger values for PSNR and SSIM are better.
    Quality was computed with
    decoded, resized, and JPEG transcoded video frames
    to decoded and resized original video frames as reference.
    }
    \label{fig:acc_jpeg_compression_image_score}

\end{figure}

%% file: fig/val_acc_ffmpeg.tex
\begin{figure}[t]
    \centering
    \def\myimgwidth{0.48}

    \centering

    \begin{minipage}{\linewidth}
    \includegraphics[width=\myimgwidth\linewidth]{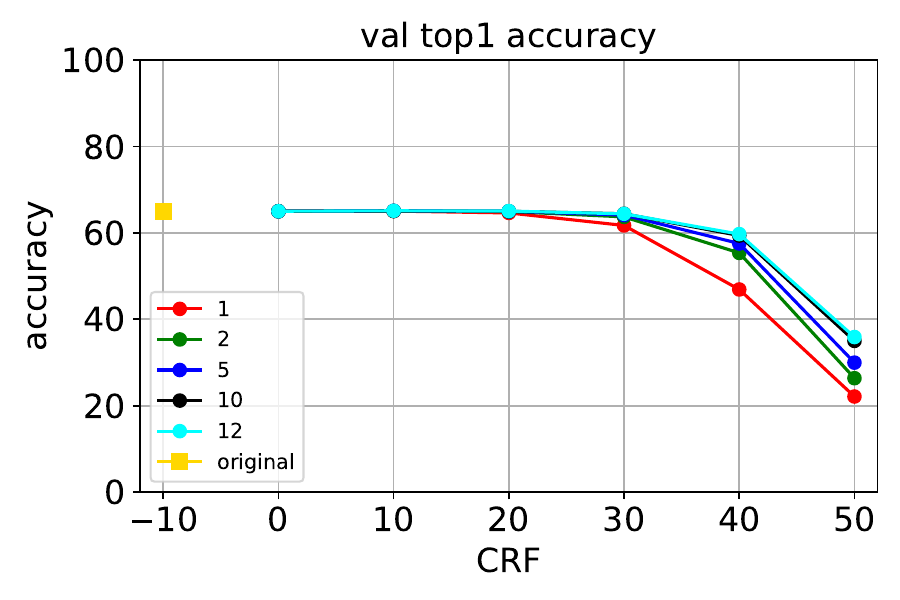}
    \includegraphics[width=\myimgwidth\linewidth]{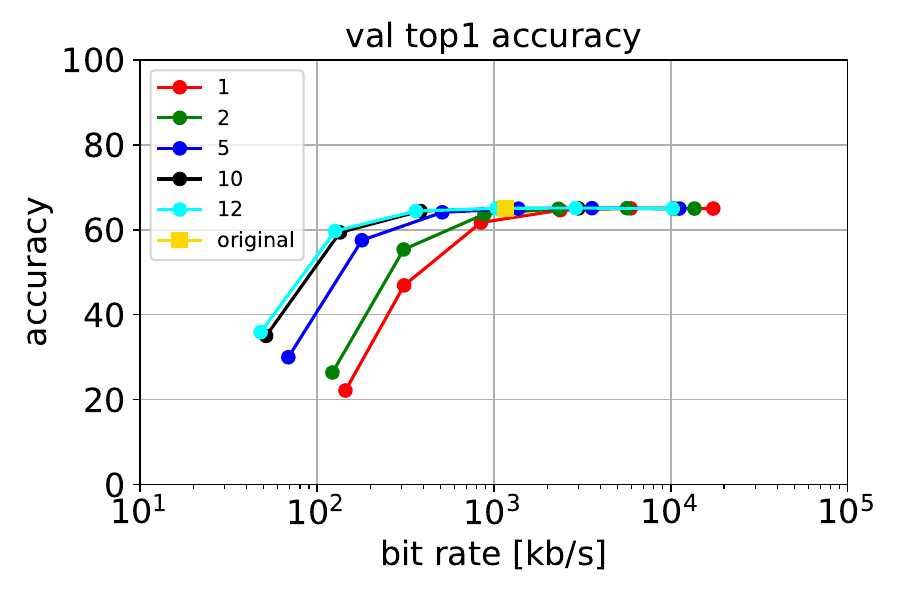}
    \vspace{-0.4cm}
    \subcaption{X3D-M}
    \label{fig:x3d_m_top1_crf}
    \label{fig:x3d_m_top1_bitrate}
    \end{minipage}
    
    \begin{minipage}{\linewidth}
    \includegraphics[width=\myimgwidth\linewidth]{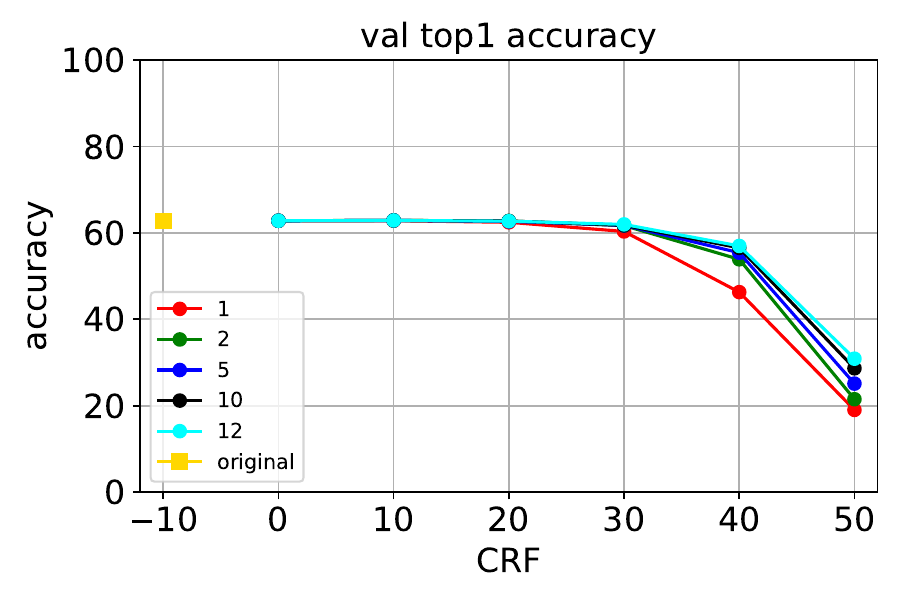}
    \includegraphics[width=\myimgwidth\linewidth]{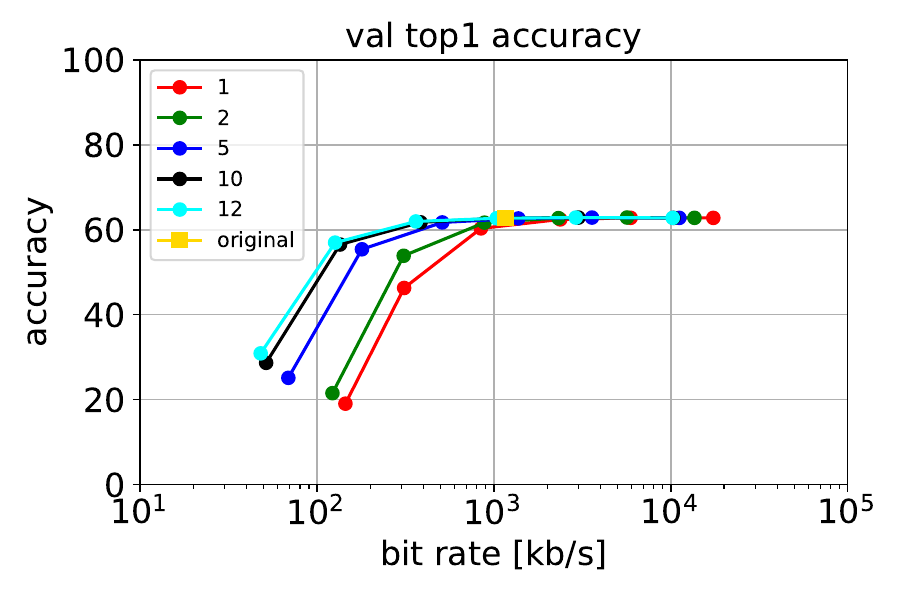}
    \vspace{-0.4cm}
    \subcaption{\modify{3D ResNet R50}}
    \label{fig:slow_r50_top1_crf}
    \label{fig:slow_r50_top1_bitrate}
    \end{minipage}

    \begin{minipage}{\linewidth}
    \includegraphics[width=\myimgwidth\linewidth]{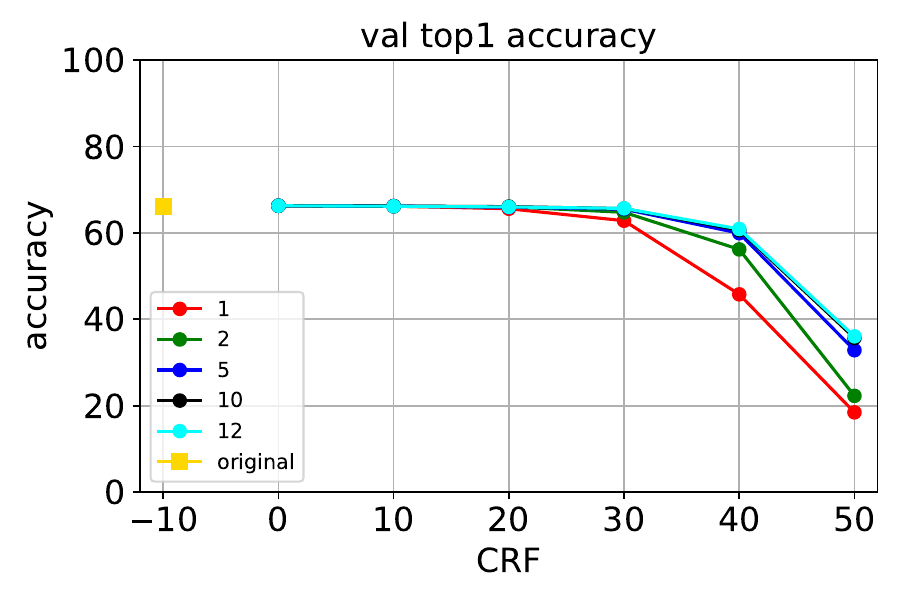}
    \includegraphics[width=\myimgwidth\linewidth]{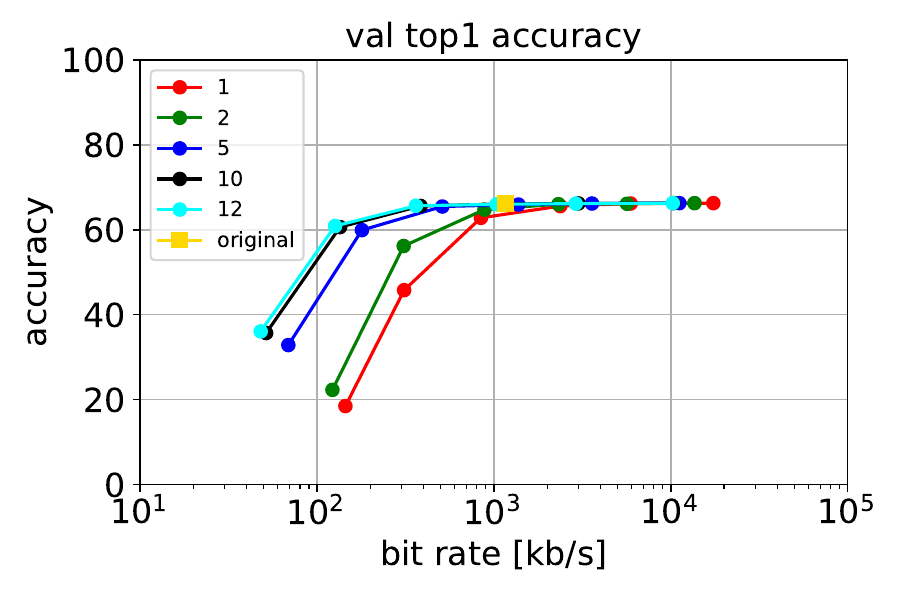}
    \vspace{-0.4cm}
    \subcaption{\modify{SlowFast R101}}
    \label{fig:slowfast_r101_top1_crf}
    \label{fig:slowfast_r101_top1_bitrate}
    \end{minipage}
    
    \begin{minipage}{\linewidth}
    \includegraphics[width=\myimgwidth\linewidth]{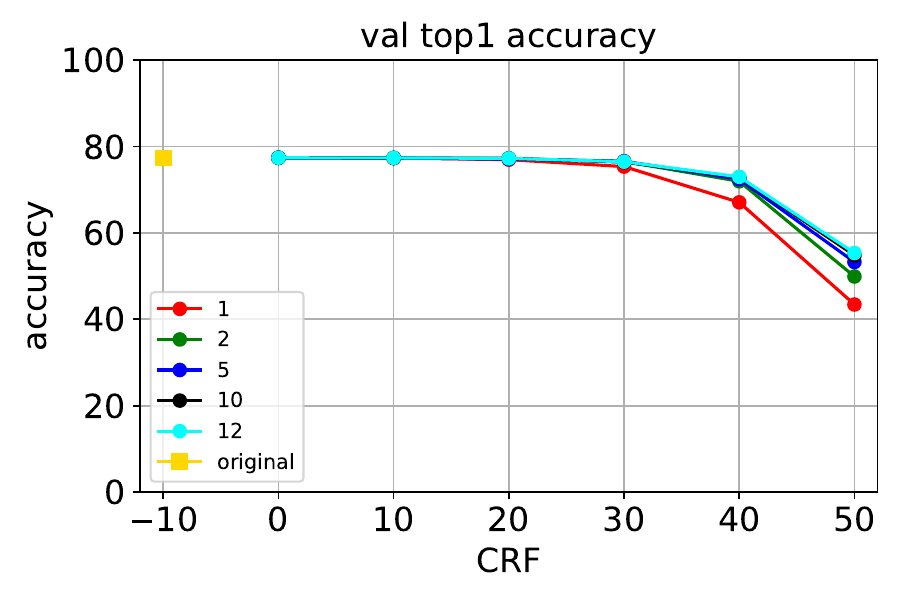}
    \includegraphics[width=\myimgwidth\linewidth]{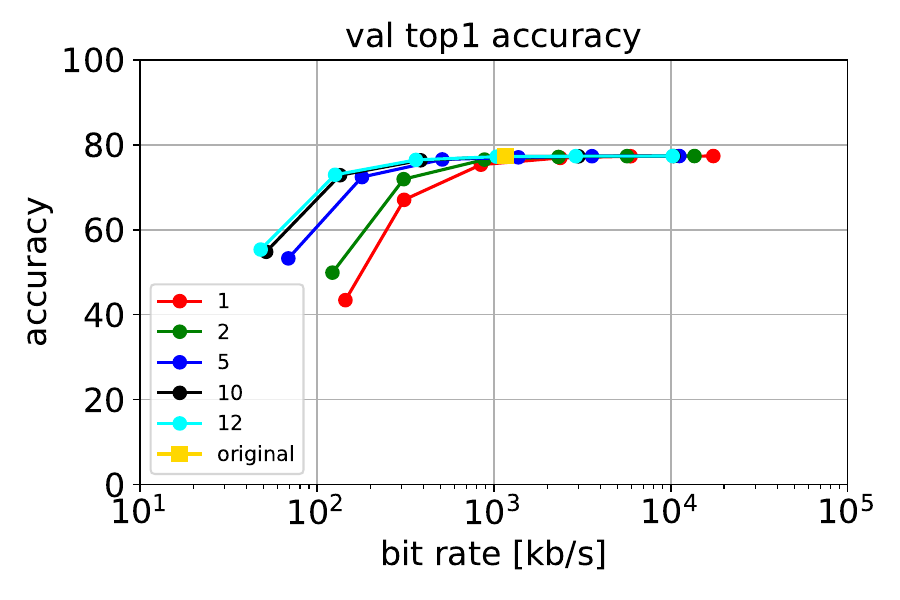}
    \vspace{-0.4cm}
    \subcaption{\modify{TimeSformer}}
    \label{fig:timesformer_top1_crf}
    \label{fig:timesformer_top1_bitrate}
    \end{minipage}
    
    \begin{minipage}{\linewidth}
    \includegraphics[width=\myimgwidth\linewidth]{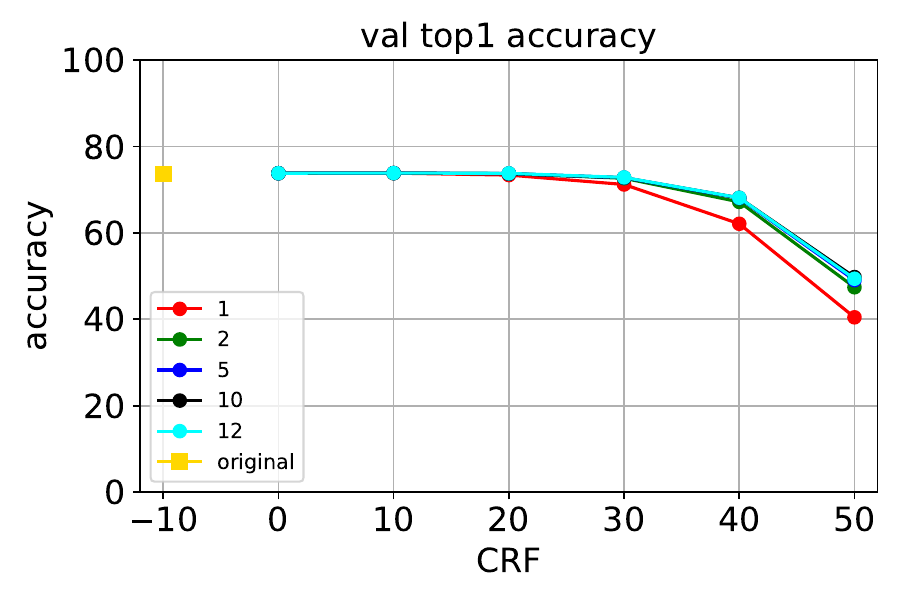}
    \includegraphics[width=\myimgwidth\linewidth]{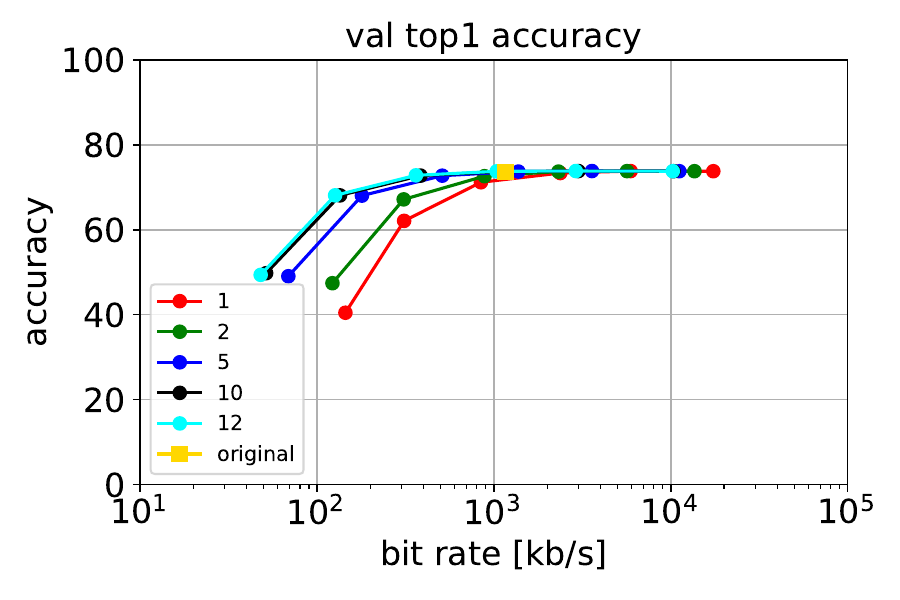}
    \vspace{-0.4cm}
    \subcaption{\modify{Video Swin-B}}
    \label{fig:videoswin_top1_crf}
    \label{fig:videoswin_top1_bitrate}
    \end{minipage}

    \caption{Top-1 Performance for videos transcoded by H.264/AVC.
    The performances of models are shown for different CRF values on the left
    (values at CRF $-10$ indicate the original videos),
    and for bit rates on the right.
    }
    \label{fig:val_acc_ffmpeg}
\end{figure}

%% file: tables/acc_ffmpeg_tables.tex
\definecolor{light_gray}{rgb}{0.9,0.9,0.9}
\newcommand{\g}{\cellcolor{light_gray}}

\begin{table}[t]

    \caption{Top-1 performance for videos transcoded by H.264/AVC.
    The values in parentheses represent the difference from the reference performance (shown on the side of model names) for the original videos.
    Shaded cells indicate the performance drops of more than 1.0\%.
    }
    \label{tab:acc_ffmpeg_x3dm}
    \label{tab:acc_ffmpeg_slowr50}
    \label{tab:acc_ffmpeg_slowfastr101}
    \label{tab:acc_ffmpeg_timesformer}
    \label{tab:acc_ffmpeg_videoswin}

    \centering

    \begin{minipage}{\linewidth}\centering
    \scalebox{0.7}{
\begin{tabular}{c|c|c|c|c|c|c}
\diagbox{GOP}{CRF}  & 0         & 10        & 20        & 30            & 40             & 50            \\ \hline
\multirow{2}{*}{1}  & 64.99     & 65.03     & 64.60     & \g  61.70     & \g  46.89      & \g  22.11     \\
                    & ($+0.07$) & ($+0.11$) & ($-0.32$) & \g  ($-3.22$) & \g  ($-18.03$) & \g ($-42.81$) \\ \hline
\multirow{2}{*}{2}  & 64.99     & 65.06     & 64.93     & \g  63.68     & \g  55.35      & \g  26.35     \\
                    & ($+0.07$) & ($+0.14$) & ($-0.01$) & \g  ($-1.24$) & \g  ($-9.57$)  & \g ($-38.57$) \\ \hline
\multirow{2}{*}{5}  & 64.99     & 65.08     & 64.97     & 64.10         & \g  57.53      & \g  29.95     \\
                    & ($+0.07$) & ($+0.16$) & ($+0.05$) & ($-0.82$)     & \g  ($-7.57$)  & \g ($-34.97$) \\ \hline
\multirow{2}{*}{10} & 64.99     & 65.04     & 64.98     & 64.41         & \g  59.37      & \g  34.98     \\
                    & ($+0.07$) & ($+0.12$) & ($+0.06$) & ($-0.51$)     & \g  ($-5.55$)  & \g ($-29.94$) \\ \hline
\multirow{2}{*}{12} & 64.99     & 65.10     & 65.03     & 64.39         & \g  59.70      & \g  35.88     \\
                    & ($+0.07$) & ($+0.18$) & ($+0.11$) & ($-0.53$)     & \g  ($-5.22$)  & \g ($-29.04$)
\end{tabular}
    }
    \subcaption{X3D-M (64.92\%)}
    \end{minipage}

    \begin{minipage}{\linewidth}\centering
    \scalebox{0.7}{
\begin{tabular}{c|c|c|c|c|c|c}
\diagbox{GOP}{CRF}  & 0         & 10        & 20          & 30           & 40            & 50             \\ \hline
\multirow{2}{*}{1}  & 62.79     & 62.80     & 62.43       & \g 60.31     & \g 46.27      & \g  19.01      \\
                    & ($+0.05$) & ($+0.06$) & ($-0.31$)   & \g ($-2.43$) & \g ($-16.47$) & \g  ($-43.73$) \\ \hline
\multirow{2}{*}{2}  & 62.80     & 62.88     & 62.74       & \g 61.68     & \g 53.86      & \g  21.49      \\
                    & ($+0.06$) & ($+0.14$) & ($\pm0.00$) & \g ($-1.06$) & \g ($-8.88$)  & \g  ($-41.25$) \\ \hline
\multirow{2}{*}{5}  & 62.79     & 62.87     & 62.66       & \g 61.72     & \g 55.39      & \g  25.10      \\
                    & ($+0.05$) & ($+0.13$) & ($-0.08$)   & \g ($-1.02$) & \g ($-7.35$)  & \g  ($-37.64$) \\ \hline
\multirow{2}{*}{10} & 62.79     & 62.87     & 62.74       & 61.75        & \g 56.51      & \g  28.64      \\
                    & ($+0.05$) & ($+0.13$) & ($\pm0.00$) & ($-0.99$)    & \g ($-6.23$)  & \g  ($-34.1$)  \\ \hline
\multirow{2}{*}{12} & 62.79     & 62.86     & 62.70       & 61.95        & \g 56.96      & \g  30.87      \\
                    & ($+0.05$) & ($+0.12$) & ($-0.04$)   & ($-0.79$)    & \g ($-5.78$)  & \g  ($-31.87$)
\end{tabular}
    }
    \subcaption{\modify{3D ResNet R50} (62.74\%)}
    \end{minipage}
    
    \begin{minipage}{\linewidth}\centering
    \scalebox{0.7}{
\begin{tabular}{c|c|c|c|c|c|c}
\diagbox{GOP}{CRF}  & 0         & 10          & 20        & 30           & 40            & 50            \\ \hline
\multirow{2}{*}{1}  & 66.26     & 66.17       & 65.57     & \g 62.80     & \g 45.75      & \g  18.45     \\
                    & ($+0.11$) & ($+0.02$)   & ($-0.58$) & \g ($-3.35$) & \g ($-20.4$)  & \g ($-47.7$)  \\ \hline
\multirow{2}{*}{2}  & 66.26     & 66.12       & 66.02     & \g 64.74     & \g 56.16      & \g 22.26      \\
                    & ($+0.11$) & ($+0.03$)   & ($-0.13$) & \g ($-1.41$) & \g  ($-9.99$) & \g ($-43.89$) \\ \hline
\multirow{2}{*}{5}  & 66.26     & 66.18       & 65.93     & 65.47        & \g 59.91      & \g  32.82     \\
                    & ($+0.11$) & ($+0.3$)    & ($-0.22$) & ($-0.68$)    & \g ($-6.24$)  & \g ($-33.33$) \\ \hline
\multirow{2}{*}{10} & 66.26     & 66.20       & 66.06     & 65.60        & \g 60.62      & \g 35.66      \\
                    & ($+0.11$) & ($+0.05$)   & ($-0.09$) & ($-0.55$)    & \g  ($-5.53$) & \g ($-30.49$) \\ \hline
\multirow{2}{*}{12} & 66.26     & 66.15       & 65.99     & 65.65        & \g 60.89      & \g  36.04     \\
                    & ($+0.11$) & ($\pm0.00$) & ($-0.16$) & ($-0.50$)    & \g ($-5.26$)  & \g ($-30.11$)
\end{tabular}
    }
    \subcaption{\modify{SlowFast R101} (66.15\%)}
    \end{minipage}
    
    \begin{minipage}{\linewidth}\centering
    \scalebox{0.7}{
\begin{tabular}{c|c|c|c|c|c|c}
\diagbox{GOP}{CRF}  & 0         & 10        & 20        & 30           & 40            & 50           \\ \hline
\multirow{2}{*}{1}  & 77.37     & 77.32     & 76.93     & \g 75.33     & \g 67.05      & \g 43.41     \\
                    & ($+0.02$) & ($-0.03$) & ($-0.42$) & \g ($-2.02$) & \g ($-10.3$)  & \g ($-33.9$) \\ \hline
\multirow{2}{*}{2}  & 77.37     & 77.34     & 77.15     & 76.53        & \g 71.92      & \g 49.89     \\
                    & ($+0.02$) & ($-0.01$) & ($-0.20$) & ($-0.82$)    & \g ($-5.43$)  & \g ($-27.5$) \\ \hline
\multirow{2}{*}{5}  & 77.37     & 77.33     & 77.06     & 76.56        & \g 72.41      & \g 53.25     \\
                    & ($+0.02$) & ($-0.02$) & ($-0.29$) & ($-0.79$)    & \g ($-4.94$)  & \g ($-24.1$) \\ \hline
\multirow{2}{*}{10} & 77.37     & 77.30     & 77.25     & 76.37        & \g 72.85      & \g 54.81     \\
                    & ($+0.02$) & ($-0.05$) & ($-0.10$) & ($-0.98$)    & \g ($-4.50$)  & \g ($-22.5$) \\ \hline
\multirow{2}{*}{12} & 77.37     & 77.34     & 77.22     & 76.46        & \g 72.95      & \g 55.33     \\
                    & ($+0.02$) & ($-0.01$) & ($-0.12$) & ($-0.89$)    & \g  ($-4.46$) & \g ($-22.0$)
\end{tabular}
    }
    \subcaption{\modify{TimeSformer} (77.35\%)}
    \end{minipage}
    
    \begin{minipage}{\linewidth}\centering
    \scalebox{0.7}{
    \begin{tabular}{c|c|c|c|c|c|c}
\diagbox{GOP}{CRF}  & 0         & 10        & 20        & 30           & 40            & 50           \\ \hline
\multirow{2}{*}{1}  & 73.79     & 73.82     & 73.36     & \g 71.19     & \g 62.09      & \g 40.45     \\
                    & ($+0.17$) & ($+0.20$) & ($-0.26$) & \g ($-2.43$) & \g ($-11.5$)  & \g ($-33.2$) \\ \hline
\multirow{2}{*}{2}  & 73.79     & 73.81     & 73.71     & 72.63        & \g 67.15      & \g 47.40     \\
                    & ($+0.17$) & ($+0.19$) & ($+0.09$) & ($-0.99$)    & \g ($-6.47$)  & \g ($-26.2$) \\ \hline
\multirow{2}{*}{5}  & 73.79     & 73.83     & 73.71     & 72.74        & \g 68.04      & \g 49.05     \\
                    & ($+0.02$) & ($+0.21$) & ($+0.09$) & ($-0.88$)    & \g ($-5.58$)  & \g ($-24.6$) \\ \hline
\multirow{2}{*}{10} & 73.79     & 73.82     & 73.69     & 72.76        & \g 68.09      & \g 49.76     \\
                    & ($+0.02$) & ($+0.20$) & ($+0.07$) & ($-0.86$)    & \g ($-5.53$)  & \g ($-23.9$) \\ \hline
\multirow{2}{*}{12} & 73.79     & 73.84     & 73.76     & 72.82        & \g 68.11      & \g  49.37    \\
                    & ($+0.02$) & ($+0.22$) & ($+0.14$) & ($-0.80$)    & \g  ($-5.51$) & \g ($-24.3$)
    \end{tabular}
    }
    \subcaption{\modify{Video Swin-B} (73.62\%)}
    \end{minipage}

\end{table}

%% file: fig/accuracy_for_each_class_x3dm.tex
\begin{figure}[t]
    \centering
    \def\myimgwidth{0.8}

    \begin{minipage}[t]{\myimgwidth\linewidth}
        \centering
        \includegraphics[width=\linewidth]{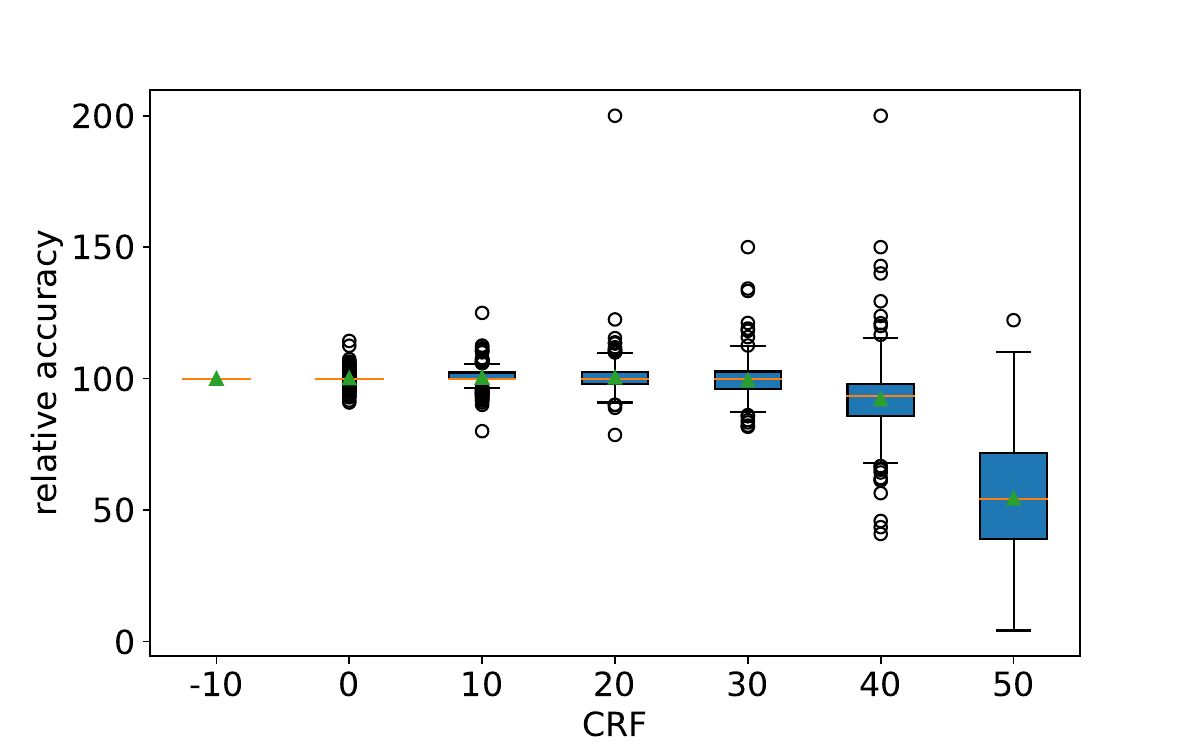} 
    \end{minipage}

    \caption{Relative top-1 performance of each class
    with different CRF values.
    The performance of the original videos (shown at CRF $-10$) is shown
    as the reference (100), and the relative performances for 400 classes are plotted as circles.
    The model was X3D-M and the GOP size was fixed to 12.
    }
    \label{fig:accuracy_for_each_class_x3dm}

\end{figure}